\documentclass[twoside]{article}
\usepackage[hyphens]{url}
\usepackage{hyperref}
\usepackage[hyphenbreaks]{breakurl}
\usepackage{xurl}
\usepackage[accepted]{aistats2021}
\usepackage[utf8]{inputenc} %
\usepackage[T1]{fontenc}    %
\usepackage{hyperref}       %
\usepackage{url}            %
\usepackage{booktabs}       %
\usepackage{amsfonts}       %
\usepackage{nicefrac}       %
\usepackage{microtype}      %
\usepackage{multirow}
\usepackage{amsmath, amssymb}
\usepackage{cleveref}
\usepackage[dvipsnames]{xcolor}
\usepackage{natbib}
\usepackage{bm}
\usepackage{graphicx}
\usepackage{sidecap}
\usepackage[toc,page]{appendix}
\usepackage{float}
\usepackage{amsthm}
\usepackage{breqn}
\hypersetup{%
    pdfpagemode={UseOutlines},
    bookmarksopen,
    pdfstartview={FitH},
    colorlinks,
    linkcolor={blue},
    citecolor={blue},
    urlcolor={black}
  }%

\def\de{\mathrm{d}}

\def\EE{\mathbb{E}}

\def\RE{\mathbb{R}}
\def\VV{\mathcal{V}}

\def\ind{\bm{1}}

\def\DP{\mathrm{DP}}

\allowdisplaybreaks

\newtheorem{theorem}{Theorem}

\newtheorem{proposition}[theorem]{Proposition}

\begin{document}

\twocolumn[

\aistatstitle{A Bayesian nonparametric approach to count-min sketch under power-law data streams}

\aistatsauthor{ Emanuele Dolera \And Stefano Favaro \And  Stefano Peluchetti }

\aistatsaddress{ emanuele.dolera@unipv.it\\ University of Pavia \And  stefano.favaro@unito.it\\University of Torino and Collegio Carlo Alberto \And speluchetti@cogent.co.jp\\Cogent Labs } ]

\begin{abstract}
The count-min sketch (CMS) is a randomized data structure that provides estimates of tokens' frequencies in a large data stream using a compressed representation of the data by random hashing. In this paper, we rely on a recent Bayesian nonparametric (BNP) view on the CMS to develop a novel learning-augmented CMS under power-law data streams. We assume that tokens in the stream are drawn from an unknown discrete distribution, which is endowed with a normalized inverse Gaussian process (NIGP) prior. Then, using distributional properties of the NIGP, we compute the posterior distribution of a token's frequency in the stream, given the hashed data, and in turn corresponding BNP estimates. Applications to synthetic and real data show that our approach achieves a remarkable performance in the estimation of low-frequency tokens. This is known to be a desirable feature in the context of natural language processing, where it is indeed common in the context of the power-law behaviour of the data. 
\end{abstract}

\section{Introduction} \label{sec:intro}

When processing large data streams of data, it is critical to represent the data in compact structures that allow to efficiently extract statistical information. Sketching algorithms, or simply sketches, are randomized data structures that can be easily updated and queried to perform a time and memory efficient estimation of statistics of large data streams of tokens. Sketches have found numerous applications in, e.g., machine learning \citep{Agg10}, security analysis \citep{Dwo10}, natural language processing \citep{Goy09}, computational biology \citep{Zha14}, social networks \citep{Son09} and games \citep{Har10}.  Of particular interest is the problem of estimating the unknown frequency of a token in a stream, which is typically referred to as a ``point query". A notable approach to address point queries is the count-min sketch (CMS) \citep{cormode2005summarizing,cor052}, which uses random hashing to obtain a compressed, or approximated, representation of tokens' frequencies in the stream. The CMS achieves the goal of using a memory efficient representation of the data stream, while having provable theoretical guarantees on the estimated point query via hashed frequencies. In the recent years, there has been an increasing interest in improving the performance of the CMS by means of learning models that allow to better exploit properties of the data. \citep{cai2018bayesian,aamand2019learned,hsu2019learning}. In this paper, we focus on the learning-augmented CMS of \citep{cai2018bayesian}, which relies on Bayesian nonparametric (BNP) modeling of a data stream of tokens.

The learning-augmented CMS of \cite{cai2018bayesian} assumes that tokens in a stream are modeled as random samples from an unknown discrete distribution, which is endowed with a Dirichlet process (DP) prior \citep{ferguson1973bayesian}. Under this BNP framework, the predictive distribution induced by the DP provides a natural generative scheme for tokens. That is, predictive distributions, combined with both a restriction property and a finite-dimensional projective property of the DP, lead to the posterior distribution of a point query, given the hashed frequencies. This is referred to as the CMS-DP. Interestingly, the posterior mode recovers the CMS estimate of \cite{cor052}, while other CMS-DP estimates, e.g. posterior mean and median, may be viewed as CMS estimates with shrinkage. The CMS-DP improves over the CMS on several aspects: i) it incorporates a priori knowledge on the data into the estimates; ii) it assumes an unknown and unbounded number of distinct tokens, which is typically expected in large datasets; iii) it allows to model, via the posterior distribution induced by a point query, the uncertainty induced by the process of random hashing.

We extend the BNP approach of \cite{cai2018bayesian} to develop a novel learning-augmented CMS under power-law data streams. Power-law distributions occur in many situations of scientific interest, and have significant consequences for the understanding of natural and man-made phenomena \citep{cla09}. Here, we assume that tokens in the stream are modeled as random samples from an unknown discrete distribution, which is endowed with a normalized inverse Gaussian process (NIGP) prior \citep{Pru02,lijoi2005hierarchical}. The NIGP comes as a ``forced" choice since it is the sole discrete nonparametric prior that combines: i) a power-law tail behaviour, in contrast with the exponential tail behaviour of the DP; ii) both a restriction property and a finite-dimensional projective property analogous to those of the DP, which are critical to compute and work with the posterior distribution of a point query given the hashed frequencies. While this prior choice limits the flexibility of tuning the prior to the power-law degree of the data, the NIGP is arguably still a sensible choice of practical interest. Under the NIGP prior, we compute the posterior distribution of a point query, given the stored hashed frequencies, and in turn corresponding BNP estimates. Applications to synthetic and real data show that our approach outperforms the CMS and the CMS-DP in the estimation of low-frequency tokens. This is known to be a desirable feature in the context of natural language processing \citep{Goy12,Goy09,Pit15}, where it is indeed common the power-law behaviour of the data stream.

The paper is structured as follows. In \Cref{sec:cms} we review the BNP approach to CMS of \cite{cai2018bayesian}, and in \Cref{sec:bayes_cms} we extend this approach to develop a novel learning-augmented CMS under power-law data streams. \Cref{sec:exp} contains numerical experiments, whereas \Cref{sec:dis} concludes with final remarks and future works.

\section{A BNP approach to CMS} \label{sec:cms}

To introduce the BNP approach of \cite{cai2018bayesian}, let $X_{1:m}=(X_{1},\ldots,X_{m})$ be a large data stream of tokens taking values in a (possibly infinite) measurable space of symbols $\VV$. The stream $X_{1:m}$ is available for inferential purposes only through its compressed representation obtained by means of random hashing. Specifically, let $J$ and $N$ be positive integers such that $[J]=\{1,\ldots,J\}$ and $[N]=\{1,\ldots,N\}$, and let $h_{1},\ldots,h_{N}$, with $h_{n}:\VV\rightarrow [J]$, be a collection of hash functions drawn uniformly at random from a pairwise independent hash family $\mathcal{H}$. For mathematical convenience it is assumed that $\mathcal{H}$ is a perfectly random hash family, that is for $h_{n}$ drawn uniformly at random from $\mathcal{H}$ the random variables $(h_{n}(x))_{x\in\mathcal{V}}$ are i.i.d. as a Uniform distribution over $[J]$. In practice, as discussed in \cite{cai2018bayesian}, real-world hash functions yield only small perturbations from perfect hash functions. Hashing $X_{1:m}$ through $h_{1},\ldots,h_{N}$ creates $N$ vectors of $J$ buckets $\{\mathbf{C}_{n}\}_{n\in[N]}$, where $\mathbf{C}_{n}=(C_{n,1},\ldots,C_{n,J})$ with $C_{n,j}$ obtained by aggregating the frequencies for all $x$ such that $h_{n}(x)=j$. Every $C_{n,j}$ is initialized at zero, and whenever a new token $X_i$ is observed we set $C_{n,h_n(X_i)} \leftarrow 1+ C_{n,h_n(X_i)}$ for every $n\in [N]$. Under this setting, the goal consists in estimating the frequency $f_{v}$ of a token of type $v\in\mathcal{V}$ in $X_{1:m}$, i.e. the point query $f_v=\sum_{1\leq i\leq m} \ind_{X_{i}}(v)$. In particular, the CMS of \cite{cor052} estimates $f_{v}$ with
\begin{equation}\label{or_estim}
\hat{f}_{v}^{\text{\tiny{(CMS)}}}=\min_{n\in[N]}\{C_{n,h_n(v)}\}_{n\in [N]}.
\end{equation}
We refer to Appendix A for a detailed account on the CMS and a theoretical (probabilistic) guarantee for the estimator \eqref{or_estim}.

Differently from the CMS of \cite{cor052}, the CMS-DP of \cite{cai2018bayesian} estimates $f_{v}$ by relying on the following modeling assumptions on the data stream $X_{1:m}$: i) symbols $v_{j}$'s in $\mathcal{V}$ are distributed as an unknown probability measure $P(\cdot) = \sum_{j\geq1} p_j \delta_{v_j}(\cdot)$ on $\mathcal{V}$; ii) $P$ is distributed as a DP prior \citep{ferguson1973bayesian} with diffuse probability (base) measure $\nu$ on $\VV$ and mass parameter $\alpha>0$. Then, tokens $X_{i}$'s are modeled as random samples from a DP, i.e., 
    \begin{align}\label{model_do}
    X_{1:m} \mid P  &\,\stackrel{\mbox{\scriptsize{iid}}}{\sim}\,P \notag\\[0.2cm]
     P &\, \sim\,\DP(\alpha,\nu)
    \end{align}
for $m\geq1$. Under \eqref{model_do}, a point query induces the posterior distribution of $f_{v}$, given $\{C_{n, h_{n}(v)}\}_{n\in[N]}$, for $v\in\VV$. CMS-DP estimates of $f_{v}$ are obtained as functionals of the posterior distribution, e.g. mode, mean, median. The computation of the posterior distribution of $f_{v}$  relies on the predictive distribution of the DP prior, namely the conditional distribution of an additional token given the stream of tokens. This is combined with two critical properties of the DP: P1) the restriction property which, due to the perfectly random $\mathcal{H}$, implies that the prior governing the tokens hashed in each of the $J$ buckets is a DP prior with mass parameter $\alpha/J$; P2) the finite-dimensional projective property which, due to the perfectly random $\mathcal{H}$, implies that the prior governing the multinomial hashed frequencies $\mathbf{C}_{n}$ is a $J$-dimensional symmetric Dirichlet distribution with parameter $\alpha/J$.

Now, we outline the BNP approach of  \cite{cai2018bayesian} based properties P1) and P2). Because of the discreteness of $P\sim\DP(\alpha,\nu)$, a random sample $X_{1:m}$ from $P$ induces a random partition of $\{1,\ldots,m\}$ into subsets labelled by distinct symbols in $\VV$. See Appendix C. The predictive distribution of the DP provides the conditional distribution, given $X_{1:m}$, over which partition subset a new token $X_{m+1}$ will join; the size of that subset is precisely the frequency $f_{v}$ we seek to estimate. However, since we have only access to the hashed frequencies, the object of interest is the distribution $p_{f_{v}}(m,\alpha)$ of $f_{v}$. This distribution follows by marginalizing out the sampling information $X_{1:,m}$, with respect to the DP prior, from the conditional distribution of $f_{v}$ given $X_{1,m}$. According to property P1), for a single $h_{n}$ the distribution $p_{f_{v}}(\cdot;c_{n, h_{n}(v)},\alpha/J)$ coincides with the posterior distribution of $f_{v}$, given $C_{n, h_{n}(v)}=c_{n, h_{n}(v)}$; the posterior distribution of $f_{v}$ given $\{C_{n, h_{n}(v)}\}_{n\in[N]}$ follows by the independence assumption on $\mathcal{H}$ and Bayes theorem. To conclude, it remains to estimate the prior's parameter $\alpha>0$ based on the hashed frequencies. According to property P2), and by the independence assumption on $\mathcal{H}$, the $N$ vectors $\{\mathbf{C}_{n}\}_{n\in[N]}$ are i.i.d. as a Dirichlet-Multinomial distribution with symmetric parameter $\alpha/J$. This fact provides an explicit expression for the likelihood function of the hashed frequencies, and thus to a Bayesian estimation of $\alpha$. 

\section{A BNP approach to CMS under power-law data streams} \label{sec:bayes_cms}

We extend the BNP approach of \cite{cai2018bayesian} to develop a novel learning-augmented CMS under power-law data streams. In this respect, it is natural to assume that tokens in a stream $X_{1:m}$ are modeled as random samples from an unknown discrete distribution $P$, and then to endow $P$ with prior distribution $\mathcal{Q}$ with power-law tail behaviour. Critical constraints in the choice of $\mathcal{Q}$ arises directly from the approach of \cite{cai2018bayesian}. In particular, the prior $\mathcal{Q}$ must feature both a restriction property and a finite-dimensional projective property analogous to those of the DP prior. This is required to compute and work with the posterior distribution of a point query, given the hashed data stream $X_{1:m}$. To the best of our knowledge, the NIGP prior \citep{Pru02,lijoi2005hierarchical} is the sole discrete nonparametric prior with power-law tail behaviour that features both a restriction property and a finite-dimensional projective property analogous to those of the DP. This paves the way to our learning-augmented CMS under data streams with power-law behaviour.

\subsection{NIGP priors}\label{sec:bayes_cms1}

The DP and the NIGP are discrete random probability measures belonging to the class of homogeneous normalized completely random measures (hNCRMs) \citep{Jam02, Pru02,Reg03, pitman2006combinatorial,lij10}. Let the measurable space $\VV$ be endowed with its Borel $\sigma$-field $\mathcal{F}$. A completely random measure CRM $\mu$ on $\VV$ is defined as a random measure such that for any $A_1,\ldots, A_k$ in $\mathcal{F}$, with $A_{i}\cap A_{j}=\emptyset$ for $i\neq j$, the random variables $\mu(A_1),\ldots,\mu(A_k)$ are mutually independent \citep{kingman2005poisson}. Any CRM $\mu$ with no fixed point of discontinuity and no deterministic drift is represented as $\mu=\sum_{j\geq1}\xi_{j}\delta_{v_{j}}$, where the $\xi_{j}$'s are positive random jumps and the $v_{j}$'s are $\VV$-valued random locations. Then, $\mu$ is characterized by the L\'evy--Khintchine representation
\begin{equation}\label{eq:levyK}
\EE\left[\text{e}^{-\int_{\VV}f(v)\mu(\text{d}v)}\right]=\text{e}^{-\int_{\mathbb{R}^{+}\times\VV}[1-\text{e}^{-\xi f(v)}]\gamma(\text{d}\xi,\text{d}v)},
\end{equation}
where $f:\VV\rightarrow\mathbb{R}$ is a measurable function such that $\int|f|\text{d}\mu<+\infty$ and $\gamma$ is a measure on $\RE^{+} \times \VV$ such that $\int_{B}\int_{\mathbb{R}^{+}}\min\{\xi,1\}\gamma(\text{d}\xi,\text{d}v)<+\infty$ for any $B\in\mathcal{F}$. For our purposes it is useful to separate the jump and location part of the L{\'e}vy intensity measure $\gamma$ by writing it as $\gamma(\text{d}\xi,\text{d}v)=\rho(\text{d}\xi;v)\nu(\text{d}v)$, where $\nu$ denotes a measure on $(\mathcal{V},\mathcal{F})$ and $\rho$ denotes a transition kernel on $\mathcal{B}(\mathbb{R}^{+})\times\mathcal{V}$, with $\mathcal{B}(\mathbb{R}^{+})$ being the Borel $\sigma$-field of $\mathbb{R}^{+}$, i.e. $v\mapsto\rho(A;v)$ is $\mathcal{F}$-measurable for any $A\in\mathcal{B}(\mathbb{R}^{+})$ and $\rho(\cdot;v)$ is a measure on $(\mathbb{R}^{+},\mathcal{B}(\mathbb{R}^{+}))$ for any $v\in\mathcal{V}$. In particular, if $\rho(\cdot;v)=\rho(\cdot)$ for any $v$ then the jumps of $\mu$ are independent of their locations. In this case, the CMR $\mu$ is termed homogeneous CRM. See Appendix B. 

hNCRMs are obtained by normalizing CRMs. To define the NIGP, we first introduce the normalized generalized Gamma process (NGGP) \citep{Jam02, Pru02, lijoi2007controlling}, which is a hNCRM including both the DP and NIGP as special cases. The NGGP is useful to understand the power-law tail behaviuor featured by the NIGP, in contrast with the exponential tail behaviour of the DP, as well as predictive properties of the NIGP. A generalized Gamma process (GGP) $\mu$ on $\mathcal{V}$ is a CRM  characterized, through the L\'evy--Khintchine formula \eqref{eq:levyK}, by the L\'evy intensity measure $\gamma(\de \xi,\de v)= \rho_{\sigma}(\de \xi)\alpha \nu(\de v)$, where: i) $\alpha >0$ is the mass parameter; ii) $\nu$ is a diffuse probability (base) measure on $\VV$, governing the location part of $\mu$; iii) $\rho_{\sigma}$, with $\sigma\in[0,1)$, is a rate measure on $\RE^{+}$ governing the jump part of $\mu$ such that
\begin{align}\label{levy_ggp}
    \rho_{\sigma}(\de \xi) = \frac{2^{-(1-\sigma)}}{\Gamma(1-\sigma)} \xi^{-(1+\sigma)} e^{-\frac{\xi}{2}} \ind_{\RE^{+}}(\xi) \de \xi.
\end{align}
See Appendix B. The total mass $\mu(\VV)$ is finite (almost surely) \citep{lijoi2007controlling}, and then the NGGP is defined as
\begin{equation}\label{def_ggp}
    P =\frac{\mu}{\mu(\VV)}= \sum_{j\geq1} p_j \delta_{v_j},
\end{equation}
where $p_{j}=\xi_{j}/\mu(\VV)$ for $j\geq1$ are random probabilities such that $p_{j}\in(0,1)$ for any $j\geq1$ and $\sum_{j\geq1}p_{j}=1$ almost surely. For short, we write $P\sim\text{NGGP}(\alpha,\sigma,\nu)$. For $\sigma=0$ the GGP reduces to the Gamma process \citep{kingman2005poisson}, and hence the NGPP becomes a DP with mass parameter $\alpha/2$. The NIGP with mass parameter $\alpha>0$ is defined as the NGGP with $\sigma=1/2$ and, for short, we write $P\sim\text{NIGP}(\alpha,\nu)$. See Appendix C.

The NIGP features a restriction property analogous to that of the DP. That is, if $A\subset\mathcal{V}$ and $P_{A}$ is the random probability measure on $A$ induced by $P\sim\text{NIGP}(\alpha,\nu)$ on $\mathcal{V}$ then $P_{A}\sim\text{NIGP}(\alpha\nu(A),\nu_{A}/\nu(A))$, where $\nu_{A}$ is the projection of $\nu$ to $A$. The restriction property of the NIGP follows from the definition of the NIGP as a normalized GGP, which has a Poisson process representation admitting the Poisson coloring theorem. See Appendix B and Chapter 5 of \citep{kingman2005poisson} for details. To the best of our knowledge, hNCRM priors are the sole discrete nonparametric priors featuring the restriction property. The NIGP also features a finite-dimensional projective property analogous that of the DP. That is, if $\{B_1,\ldots,B_{k}\}$ is a measurable $k$-partition of $\VV$, for any $k\geq1$, then $P\sim \text{NIGP}(\alpha, \nu)$ is such that  
\begin{equation}\label{finite_dim}
(P(B_1), \ldots, P(B_k))\stackrel{\text{d}}{=}\left(\frac{W_{1}}{\sum_{i=1}^{k}W_{i}},\ldots,\frac{W_{k}}{\sum_{i=1}^{k}W_{i}}\right),
\end{equation}
with $\stackrel{\text{d}}{=}$ denoting an equality in distribution, where the $W_{i}$'s are independent random variables distributed as an inverse Gaussian (IG) distribution \citep{Ses93} with shape parameter $\alpha\nu(B_{i})$ and scale parameter $1$, for $i=1,\ldots,k$. The distribution of $(P(B_1), \ldots, P(B_k))$ is referred to as the normalized IG distribution \citep{lijoi2005hierarchical,had(11)} with parameter $(\alpha\nu(B_{1}),\ldots,\alpha\nu(B_{k}))$. The finite-dimensional projective property of the NIGP follows directly from the definition of the NIGP through its finite-dimensional distributions \citep{lijoi2005hierarchical}, for which it is critical a peculiar additive property of the IG distribution. See Appendix D. To the best of our knowledge, the DP prior and the NIGP prior are the sole hNCRM priors featuring the finite-dimensional projective property.

Before describing the power-law tail behaviour of the NGGP prior, and hence the power-law tail behaviour featured by the NIGP prior, we recall the sampling structure of the NGGP. Hereafter, we denote by $(a)_{(n)}$ the ascending factorial of $a$ of order $n$, i.e., $(a)_{(n)}=\prod_{0\leq i\leq n-1}(a+i)$. Let $P\sim\text{NGGP}(\alpha,\sigma,\nu)$, with $\sigma\in(0,1)$. Because of the discreteness of $P$, a random sample of tokens $X_{1:m}$ from $P$ induces a random partition of the set $\{1,\ldots,m\}$ into $1\leq K_{m}\leq m$ partition subsets, labelled by distinct symbols $\mathbf{v}=\{v_1,\ldots,v_{K_{m}}\}$, with corresponding frequencies $(N_{1},\ldots,N_{K_{m}})$ such that $1\leq N_{i}\leq n$ and $\sum_{1\leq i\leq K_{n}}N_{i}=n$. For any $1\leq r\leq m$, let $M_{r,m}\geq0$ denote the random number of distinct symbols with frequency $r$, i.e. $M_{r,m}=\sum_{1\leq i\leq K_{m}} \ind_{N_{i}}(r)$ such that $\sum_{1\leq r\leq m}M_{r,m}=K_{m}$ and $\sum_{1\leq r\leq m}rM_{r,m}=m$. The distribution of $\mathbf{M}_{m}=(M_{1,m},\ldots,M_{m,m})$ is defined on the set $\mathcal{M}_{m,k}=\{(m_{1},\ldots,m_{n})\text{ : }m_{i}\geq0,\,\sum_{1\leq i\leq m}m_{i}=k,\,\sum_{1\leq i\leq m}im_{i}=m\}$. See Appendix C. For $\mathbf{m}\in\mathcal{M}_{m,k}$
\begin{equation}\label{eq:distribution}
\text{Pr}[\mathbf{M}_{m}=\mathbf{m}]=V_{m,k}m!\prod_{i=1}^{m}\left(\frac{(1-\sigma)_{(i-1)}}{i!}\right)^{m_{i}}\frac{1}{m_{i}!},
\end{equation}
where
\begin{align*}
    V_{m,k} &=\frac{\alpha^{k} 2^{m-k}\text{e}^{\frac{\alpha}{2\sigma}}}{\Gamma(m)}\int_{0}^{+\infty}\frac{x^{m-1}\text{e}^{-\frac{\alpha}{2\sigma}\left(1+2x\right)^{\sigma}}}{(1+2x)^{m-k\sigma}}\de x.
\end{align*}
In the the next proposition we state the predictive distribution of $P\sim\text{NGGP}(\alpha,\sigma,\nu)$ as a function of the sampling information $X_{1:m}$ through the statistic $\mathbf{M}_{m}$. The predictive distribution of the DP prior arises by letting $\sigma\rightarrow0$, whereas the predictive distribution of the NIGP prior arises by setting $\sigma=1/2$. See Appendix C.

\begin{proposition}\label{discovery}
For any $m\geq1$, let $X_{1:m}$ be a random sample from $P\sim\text{NGGP}(\alpha,\sigma,\nu)$, with $\sigma\in(0,1)$, and let $X_{1:m}$ feature $K_{m}=k$ partition subsets, labelled by $\mathbf{v}=\{v_1,\ldots,v_{K_{m}}\}$, with frequencies $(N_{1},\ldots,N_{K_{m}})$ such that $M_{r,m}=m_{r}$ for $1\leq r\leq m$. Let $\mathbf{v}_{r}=\{v_{i}\in \mathbf{v}\text{ : } N_{i}=r\}$, i.e., the labels of the partition subsets with frequency $r$ in $\mathbf{v}$, and $\mathbf{v}_{0}=\VV-\mathbf{v}$, i.e., the labels in $\VV$ not in $\mathbf{v}$. Then,
\begin{align} \label{eq:pred_seen_gg}
\text{Pr}[X_{m+1} \in \mathbf{v}_{r} \mid X_{1:m}]=
\begin{cases} 
\frac{V_{m+1, k+1}}{V_{m,k}}&\quad r=0\\[0.4cm] 
\frac{V_{m+1,k}}{V_{m,k}}(r-\sigma)m_{r}&\quad r\geq1.
\end{cases}
\end{align}
\end{proposition}

\begin{figure}[!htb]
    \centering
    \includegraphics[width=\linewidth]{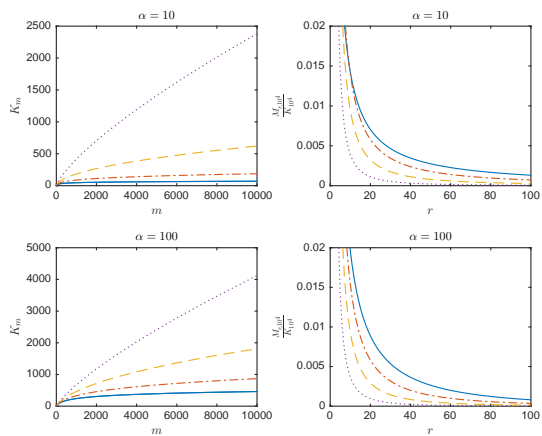}
    \caption{$K_{m}$ and $\frac{M_{r,10^{4}}}{K_{10^{4}}}$ under $P\sim\text{NGGP}(\alpha,\sigma,\nu)$: $\sigma=0$ (blue -), $\sigma=.25$ (red -.), $\sigma=.5$ (yellow --) $\sigma=.75$ (purple :)}
    \label{fig:draws}
\end{figure}

Let $P\sim\text{NGGP}(\alpha,\sigma,\nu)$ with $\sigma\in(0,1)$ and, from the definition of $P$ in \eqref{def_ggp}, let $(p_{(j)})_{i\geq}$ denote the decreasing ordered random probabilities $p_{j}$'s of $P$. By combining the rate measure \eqref{levy_ggp} with Proposition 23 of \cite{gne(07)}, as $j\rightarrow+\infty$ the $p_{(j)}$'s follow a power-law distribution of exponent $s=\sigma^{-1}$. See \cite{pit03} and references therein for details. That is, the parameter $\sigma\in(0,1)$ controls the power-law tail behaviour of $P$ through the small probabilities $p_{(j)}$'s: the larger $\sigma$ the heavier the tail of $P$. At the sampling level, the power-law behaviour of $P\sim\text{NGGP}(\alpha,\sigma,\nu)$ emerges directly from the large $m$ asymptotic behaviour of the statistics $K_{m}$ and $M_{r,m}/K_{m}$ induced by \eqref{eq:pred_seen_gg}. In particular, let $X_{1:m}$ be a random sample from $P$. Then, in Proposition 3 of \citep{lijoi2007controlling} it is showed that, as $m\rightarrow+\infty$,
\begin{align} \label{eq:sigma_diversity}
   \frac{K_{m}}{m^{\sigma}}\rightarrow S_{\sigma}
\end{align}
almost surely, where $S_{\sigma}$ is a positive and finite (almost surely) random variable \citep{pitman2006combinatorial}. Moreover, as $m\rightarrow+\infty$
\begin{align} \label{eq:sigma_diversity_l}
    \frac{M_{r,m}}{K_{m}}\rightarrow\frac{\sigma(1-\sigma)_{(r-1)}}{r!}
\end{align}
almost surely. Equation \eqref{eq:sigma_diversity} shows that the number $K_{m}$ of distinct symbols in $X_{1:m}$, for large $m$, grows as $m^{\sigma}$. This is the growth of the number of distinct symbols in random samples from a power-law distribution of exponent $s=\sigma^{-1}$. Moreover, Equation \eqref{eq:sigma_diversity_l} shows that $p_{\sigma,r}=\sigma(1-\sigma)_{(r-1)}/r!$ is the large $m$ asymptotic proportion of the number of distinct symbols with frequency $r$. Then $ p_{\sigma,r}\approx c_{\sigma}r^{-\sigma -1}$ for large $r$, for a constant $c_{\sigma}$. This is the distribution of the number of distinct symbols with frequency $r$  in random samples from a power-law distribution of exponent $s=\sigma^{-1}$. See Figure \ref{fig:draws}.

\subsection{The CMS-NIGP}\label{sec:bayes_cms2}

Because of its unicity in combining a power-law tail behaviour with both a restriction property and a finite-dimensional projective property, the NIGP prior comes as a ``forced" choice within our problem of extending the BNP approach of \cite{cai2018bayesian} to deal with power-law data streams. While this choice limits the flexibility of tuning the prior to the power-law degree of the data, in the sense that the NIGP prior is defined as a NGGP prior with $\sigma=1/2$, it is still a sensible choice of practical interest in applications. In particular, if one were forced to choose a single value for $\sigma\in(0,1)$, without information on the power-law degree of the data, $\sigma=1/2$ would arguably be a sensible and safe choice. Hereafter, we assume that tokens in a stream $X_{1:m}$ are modeled as random samples from the NIGP, i.e., 
    \begin{align*}
    X_{1:m} \mid P  &\,\stackrel{\mbox{\scriptsize{iid}}}{\sim}\,P \notag\\[0.2cm]
     P &\,\sim\,\text{NIGP}(\alpha,\nu)
    \end{align*}
for $m\geq1$. Tokens $X_{i}$'s are hashed through a collection of hash functions $h_{1},\ldots,h_{N}$ drawn uniformly at random from a pairwise independent hash family $\mathcal{H}$ which, for mathematical convenience, it is assumed to be perfectly random. Under this BNP setting, we combine the predictive distribution of the NIGP with both its restriction property and the finite-dimensional projective property to develop a learning-augmented CMS under power-law data streams. This is referred to as the CMS-NIGP. In particular, we show that, a point query induces the posterior distribution for the frequency $f_{v}$ of a token of type $v$ in $X_{1:m}$, given the hashed frequencies $\{C_{n, h_{n}(v)}\}_{n\in[N]}$, for $v\in\VV$. CMS-NIGP estimates of $f_{v}$ are obtained as suitable functionals of the posterior distribution, e.g. mode, mean, median.

The predictive distribution of $P\sim\text{NIGP}(\alpha,\nu)$,  i.e. Equation \eqref{eq:pred_seen_gg} with $\sigma=1/2$, induces the conditional distribution of $f_{v}$, given $\mathbf{M}_{m}$. However, since we have only access to the hashed frequencies $\{C_{n, h_{n}(v)}\}_{n\in[N]}$, the object of interest is in the distribution $p_{f_{v}}(m,\alpha)$ of $f_{v}$. Therefore $\mathbf{M}_{m}$ must be marginalized out, under the NIGP prior, from the conditional distribution of $f_{v}$ given $\mathbf{M}_{m}$. That is, the distribution of $f_{v}$ is obtained as
\begin{align*}
&p_{f_{v}}(\ell;m,\alpha)=\text{Pr}[f_{v}=\ell]\\
&\quad=\sum_{\mathbf{m}\in\mathcal{M}_{k,m}}\text{Pr}[X_{m+1} \in \mathbf{v}_{\ell} \mid \mathbf{M}_{m}=\mathbf{m}]\text{Pr}[\mathbf{M}_{m}=\mathbf{m}]
\end{align*}
for $\ell=0,1,\ldots,m$, where the predictive distribution $\text{Pr}[X_{m+1} \in \mathbf{v}_{\ell} \mid \mathbf{M}_{m}=\mathbf{m}]$ arises from \eqref{eq:pred_seen_gg} with $\sigma=1/2$, and the distribution $\text{Pr}[\mathbf{M}_{m}=\mathbf{m}]$ arises from \eqref{eq:distribution} with $\sigma=1/2$. The next proposition combines \eqref{eq:pred_seen_gg} and \eqref{eq:distribution} with $\sigma=1/2$ to compute the distribution $p_{f_{v}}(\ell;m,\alpha)$. In this respect, we exploit the fact that the predictive distribution of the NGGP is a function of simple sufficient statistics of the data stream $X_{1:m}$, i.e. the statistics $K_{n}$ and $(K_{n},M_{r,m})$ for $r=1,\ldots,m$. This peculiar feature of the NGGP \citep{bac17} prior allows to obtain a workable expression, from a purely computational perspective, of $p_{f_{v}}(\ell;m,\alpha)$. See Appendix E.

\begin{proposition}\label{prp_main}
For any $m\geq1$, let $X_{1:m}$ denote a random sample of tokens from $P\sim\text{NIGP}(\alpha,\nu)$. Then, 
\begin{align}\label{pred_marg_nig}
p_{f_{v}}(\ell;m,\alpha)=
\begin{cases} 
\frac{{m\choose \ell}e^{\alpha}\alpha}{\pi}\\
\quad\times\int_{0}^{1}K_{-1}\left(\frac{\alpha}{\sqrt{x}}\right)\frac{x^{m-\ell-1}}{(1-x)^{-\frac{1}{2}-\ell+1}}\de x\\
\quad\quad\quad\quad\quad\quad\quad\quad \ell=0,1,\ldots,m-1\\[0.4cm] 
\frac{2^{m}\alpha\left(\frac{1}{2}\right)_{(m)}}{\Gamma(m+1)}\\
\quad\times\int_{0}^{+\infty}\frac{x^{m}\text{e}^{-\alpha(\sqrt{1+2x}-1)}}{(1+2x)^{m+1/2}}\de x\\
\quad\quad\quad\quad\quad\quad\quad\quad \ell=m,
\end{cases}
\end{align}
where $K_{-1}(\cdot)$ is the modified Bessel function of the second type, or Macdonald function, with parameter $-1$. 
\end{proposition}

Uniformity of the hash function $h\sim\mathcal{H}$ implies that each hash function $h_{n}$ induces a $J$-partition of $\mathcal{V}$, say $\{B_{h_{n},1},\ldots,B_{h_{n},J}\}$, and the measure with respect to $P\sim \text{NIGP}(\alpha, \nu)$ of each $B_{h_{n},j}$ is $1/J$. By the restriction property of the NIGP, the hash function $h_{n}$ turns a global $P\sim\text{NIGP}(\alpha,\nu)$ that governs the distribution of  $X_{1:m}$ into a collection of bucket-specific $P_{j}\sim\text{NIGP}(\alpha/J,J\nu_{B_{h_{n},j}})$, for $j=1,\ldots,J$, that govern the distribution of the sole tokens that hashed there. This, combined Proposition \ref{prp_main}, leads to the posterior distribution, for the single hash function $h_{n}$, of $f_{v}$ given $C_{n, h_{n}(v)}$, i.e.,
\begin{equation}\label{post_first}
\text{Pr}[f_{v}=\ell\,|\,C_{n, h_{n}(v)}=c_{n, h_{n}(v)}]=p_{f_{v}}\left(\ell;c_{n, h_{n}(v)},\frac{\alpha}{J}\right)
\end{equation}
for $\ell=0,1,\ldots,c_{n, h_{n}(v)}$ and $n\in[N]$. Then, the posterior distribution of $f_{v}$ given $\{C_{n, h_{n}(v)}\}_{n\in[N]}$ follows from the posterior distribution \eqref{post_first} by exploiting the independence assumption of $\mathcal{H}$ and by Bayes theorem. This posterior distribution, which is reported in the next theorem, is the core of the CMS-NIGP. See Appendix E.

\begin{theorem}\label{thm_main}
Let $h_{1},\ldots,h_{N}$ be hash functions drawn at random from a truly random hash family $\mathcal{H}$. For any $m\geq1$, let $X_{1:m}$ be a random sample of tokens from $P\sim\text{NIGP}(\alpha,\nu)$ and let $\{C_{n, h_{n}(v)}\}_{n\in[N]}$ be the hashed frequencies induced form $X_{1:m}$ through $h_{1},\ldots,h_{N}$, i.e. $C_{n,h_{n}(v)} = \sum_{1\leq i\leq m} \ind_{h_{n}(X_i) }(h_{n}(v))$ for $n\in[N]$ and $v\in\VV$. Then, the posterior distribution of $f_{v}$, given $\{C_{n, h_{n}(v)}\}_{n\in[N]}$, is
\begin{align} \label{eq:final}
    &\text{Pr}[f_v = \ell \mid \{C_{n, h_{n}(v)}\}_{n\in[N]}=\{c_{n, h_{n}(v)}\}_{n\in[N]}] \\
    &\quad\notag\propto \prod_{n\in[N]} \begin{cases} 
\frac{{c_{n,h_n(v)}\choose \ell}e^{\frac{\alpha}{J}}\alpha}{J\pi}\\
\quad\times\int_{0}^{1}K_{-1}\left(\frac{\alpha}{J\sqrt{x}}\right)\frac{x^{c_{n,h_n(v)}-\ell-1}}{(1-x)^{-\frac{1}{2}-\ell+1}}\de x\\
\quad\quad\quad\quad\quad\quad\quad \ell=0,1,\ldots,c_{n,h_n(v)}-1\\[0.4cm] 
\frac{2^{c_{n,h_n(v)}}\alpha\left(\frac{1}{2}\right)_{(c_{n,h_n(v)})}}{J\Gamma(c_{n,h_n(v)}+1)}\\
\quad\times\int_{0}^{+\infty}\frac{x^{c_{n,h_n(v)}}e^{-\frac{\alpha}{J}(\sqrt{1+2x}-1)}}{(1+2x)^{c_{n,h_n(v)}+1/2}}\de x\\
\quad\quad\quad\quad\quad\quad\quad\ell=c_{n,h_n(v)},
\end{cases}
\end{align}
where $K_{-1}(\cdot)$ is the modified Bessel function of the second type, or Macdonald function, with parameter $-1$. 
\end{theorem}

 While computing the posterior distribution of $f_{v}$ is more than what is required from the classical CMS, it leads to two main advantages: i) the posterior distribution of $f_{v}$ allows to compute different CMS estimates of $f_{v}$ according to the specification of suitable loss functions, e.g. posterior mean under a quadratic loss, posterior median under the absolute loss, posterior mode under the $0$-$1$ loss; ii) the posterior distribution of $f_{v}$ provides a natural tool to quantify uncertainty of CMS estimates, e.g. via the posterior variance or, in general, via credible intervals arising from suitable concentration inequalities. With respect to i), \cite{cai2018bayesian} showed that the posterior mode recovers the CMS estimate, and they applied the posterior mean to improve CMS estimates of low-frequency tokens. In our context of power-law data streams, we will consider the posterior mean, which is shown to provide better estimates of low-frequency tokens. With respect to ii), to the best of our knowledge, there are no studies investigating the problem of assessing uncertainty of CMS estimates. The BNP approach improves over CMS's algorithms, by providing with a natural tool, i.e. the posterior distribution, for quantifying uncertainty of CMS estimates.

To conclude our posterior analysis of $f_{v}$, it remains to estimate the prior's parameter $\alpha>0$ from the collection of hashed frequencies. In particular, this  step requires the distribution of $\{\mathbf{C}_{n}\}_{n\in[N]}$, that is the likelihood function of the hashed frequencies. According to the finite-dimensional projective property of the NIGP prior, for a single hash function $h_{n}$ the distribution of the hashed frequencies $\mathbf{C}_{n}$ is obtained by integrating the normalized IG distribution \eqref{finite_dim} with parameter $(\alpha/J,\ldots,\ldots,\alpha/J)$ against the multinomial counts $(\mathbf{c}_{n})$. Then, the distribution of  $\{\mathbf{C}_{n}\}_{n\in[N]}$ follows by the independence assumption of the hash family $\mathcal{H}$. That is,
\begin{align}\label{marg}
&\text{Pr}[\{\mathbf{C}_{n}\}_{n\in[N]}=\{\mathbf{c}_{n}\}_{n\in[N]}]\\
&\notag\quad=\prod_{n\in[N]}\frac{m\left(\frac{\alpha}{J}\right)^{m+\frac{J}{2}}\text{e}^{\alpha}}{(\pi/2)^{\frac{J}{2}}\prod_{j=1}^{J}c_{n,j}!}\\
&\notag\quad\quad\times\int_{0}^{+\infty}\frac{x^{m-1}\prod_{j=1}^{J}K_{c_{n,j}-\frac{1}{2}}\left(\frac{\left(\frac{\alpha}{J}\right)}{(1+2x)^{-1/2}}\right)}{(1+2x)^{\frac{m}{2}-\frac{J}{4}}}\de x.
\end{align}
See Appendix F. Equation \eqref{marg} provides an explicit expression of the likelihood function of the hashed frequencies $\{\mathbf{c}_{n}\}_{n\in[N]}$, and thus it allows for estimating the parameter $\alpha$. Here, we adopt an empirical Bayes approach to estimate $\alpha$: we maximize, with respect to $\alpha$, the likelihood function of the hashed frequencies. Alternatively, a fully Bayesian approach can be considered by placing a  prior distribution on $\alpha$. The maximization of the likelihood function is performed by evaluating \eqref{marg} on a grid of exponentially spaced values, which gives an initial bracketing of the maximum. Then, we apply the golden section search derivative-free optimization algorithm \citep{press2007numerical} to maximize \eqref{marg} up to a given absolute tolerance for $\alpha$. See Appendix F.

\section{Experiments} \label{sec:exp}

We apply the CMS-NIGP to synthetic data and to real data. For the CMS-NIGP estimator of $f_{v}$, we consider the posterior mean $\hat{f}_{v}^{\text{\tiny{(NIGP)}}}$, which follows from: i) the maximization of \eqref{marg} with respect to the parameter $\alpha$ for the given data stream of tokens; ii) the evaluation, under the selected (optimal) $\alpha$, of \eqref{eq:final} for the given data stream of tokens. To ensure numerical stability in \texttt{float64}, it is imperative to work in log-space and employ the log-sum-exp trick for all summations, including the quadratures to evaluate integrals. In general, the function $K_{\nu}(x)$ is difficult to evaluate for an arbitrary real-valued $\nu$. Hoverer, for the special case $\nu=c_{n,j}-1/2$ considered in \eqref{marg}, $K_{c_{n,j}-1/2}(x)$ admits a finite sum representation that simplify its evaluation. See Appendix F. In \eqref{eq:final} instead $\nu$ is integer-valued and numerically accurate implementations are available (\texttt{Stan} and \texttt{Boost} \texttt{C++}). We checked that the number of employed quadrature points resulted in converged numerical estimates, and that the evaluations of \eqref{eq:final} passed basic sanity checks. The computational complexity for evaluating \eqref{marg} directly is $\mathcal{O}(NQJ)$, where $Q$ is the number of quadrature points; we reduced this quantity by caching the evaluations of $K_{c_{n,j}-1/2}$ for each step of the optimization. Instead the evaluation of \eqref{eq:final} has $\mathcal{O}(NQ\min_n{c_{n,h_n(v)}})$ computational complexity. The evaluation of \eqref{eq:final} and \eqref{marg} have been performed  on a MacBook Pro, and it takes about ten minutes.

We compare the CMS-NIGP estimator $\hat{f}_{v}^{\text{\tiny{(NIGP)}}}$ with respect to: i)  the CMS estimator $\hat{f}_{v}^{\text{\tiny{(CMS)}}}$ in \cite{cor052}, namely the minimum hashed frequency based on $N$ hash functions; ii) the CMS-DP estimator $\hat{f}_{v}^{\text{\tiny{(DP)}}}$ of \cite{cai2018bayesian} corresponding to the posterior mean under the DP prior. We also consider the count-mean-min (CMM) estimator $\hat{f}_{v}^{\text{\tiny{(CMM)}}}$ discussed in the work of \cite{Goy12}. The CMM relies on the same summary statistics, i.e. the buckets $\{\mathbf{C}_{n}\}_{n\in[N]}$, applied in the CMS, CMS-DP and CMS-NIGP. This facilitates the implementation of a fair comparison among estimators, since the storage requirement and sketch update complexity are unchanged. In particular, \cite{Goy12} shows that the CMM estimator stands out in the estimation of low-frequency tokens (see Figure 1 of \cite{Goy12}), which is a desirable feature in the context of natural language processing where it is common the power-law behaviour of the data stream of tokens. Hereafter, we compare estimators $\hat{f}_{v}^{\text{\tiny{(NIGP)}}}$, $\hat{f}_{v}^{\text{\tiny{(DP)}}}$, $\hat{f}_{v}^{\text{\tiny{(CMS)}}}$ and $\hat{f}_{v}^{\text{\tiny{(CMM)}}}$ in terms of the MAE (mean absolute error) between true frequencies and their corresponding estimates. Because of the limitation of page space, the comparison of $\hat{f}_{v}^{\text{\tiny{(NIGP)}}}$ with respect to $\hat{f}_{v}^{\text{\tiny{(CMS)}}}$ and $\hat{f}_{v}^{\text{\tiny{(CMM)}}}$ on synthetic data is reported in Appendix G. The comparison of $\hat{f}_{v}^{\text{\tiny{(NIGP)}}}$ with respect to $\hat{f}_{v}^{\text{\tiny{(CMS)}}}$ on real data is also in Appendix G.

We consider datasets of tokens simulated from Zipf's distribution with (exponent) parameter $s>1$, denoted by $\mathcal{Z}_{s}$. The parameter $s$ controls the tail behaviour of the Zipf's distribution: the smaller $s$ the heavier is the tail of the distribution, i.e., the smaller $s$ the larger the fraction of symbols with low-frequency tokens. Here, we generate synthetic datasets of $m=500.000$ tokens from a Zipf's distributions with parameter $s=1.3,\,1.6,\,1.9,\,2.2,\,2.5$. We make use of a $2$-universal hash family, with the following pairs of hashing parameters: i) $J=320$ and $N=2$; ii) $J=160$ and $N=4$. Table \ref{tab:experiment_sin} and Table \ref{tab:experiment_sin_2} reports the MAE of the estimators $\hat{f}_{v}^{\text{\tiny{(DP)}}}$ and $\hat{f}_{v}^{\text{\tiny{(NIGP)}}}$. From  Table \ref{tab:experiment_sin} and Table \ref{tab:experiment_sin_2}, it is clear that $\hat{f}_{v}^{\text{\tiny{(NIGP)}}}$ has a remarkable better performance than $\hat{f}_{v}^{\text{\tiny{(DP)}}}$ in the estimation of low-frequency tokens. In particular, for both Table \ref{tab:experiment_sin} and Table \ref{tab:experiment_sin_2}, if we consider the bin of low-frequencies $(0,32]$ the MAE of $\hat{f}_{v}^{\text{\tiny{(NIGP)}}}$ is alway smaller than the MAE of $\hat{f}_{v}^{\text{\tiny{(DP)}}}$, i.e. $\hat{f}_{v}^{\text{\tiny{(NIGP)}}}$ outperforms $\hat{f}_{v}^{\text{\tiny{(DP)}}}$. This behaviour becomes more and more evident as the parameter $s$ decreases, that is the heavier is the tail of the distribution the more the estimator $\hat{f}_{v}^{\text{\tiny{(NIGP)}}}$ outperforms the estimator $\hat{f}_{v}^{\text{\tiny{(DP)}}}$. In particular, for dataset $\mathcal{Z}_{1.5}$ the CMS-NIGP outperforms the CMS-DP for tokens with frequency smaller than 256, whereas for dataset $\mathcal{Z}_{2.5}$ the CMS-NIGP outperforms the CMS-DP for tokens with frequency smaller than 16. 

A comparison between $\hat{f}_{v}^{\text{\tiny{(NIGP)}}}$, $\hat{f}_{v}^{\text{\tiny{(CMS)}}}$ and  $\hat{f}_{v}^{\text{\tiny{(CMM)}}}$ is reported in Appendix G. This comparison reveals that the CMS-NIGP outperforms the CMS in the estimation of low-frequency tokens for both the choices of hashing parameters, whereas the CMS-NIGP outperforms the CMM in the estimation of low-frequency token for the choice of hashing parameters $J=160$ and $N=4$. In general, from our experiments it emerges that $\hat{f}_{v}^{\text{\tiny{(NIGP)}}}$ underestimates large-frequency tokens. To explain this underestimation phenomenon, we observe that the posterior distribution $p_{f_{v}}(\ell;c_{n, h_{n}(v)},\alpha/J)$ in \eqref{post_first} is a decreasing function of $\ell\in\{1,2,\ldots,m\}$. In other terms, the posterior distribution of $f_{v}$ assigns more probability mass to small values of $f_{v}$. Such a decreasing behaviour of $p_{f_{v}}(\ell;c_{n, h_{n}(v)},\alpha/J)$, which is is inherited from the predictive distribution of the NIGP prior, provides an intuitive explanation of the empirical evidence that the larger $v$ the more $\hat{f}_{v}^{\text{\tiny{(NIGP)}}}$ underestimates $f_{v}$, i.e. for any parameter $s=1.3,\,1.6,\,1.9,\,2.2,\,2.5$ of Zipf's distribution the MAE increases along the rows of Table \ref{tab:experiment_sin} and Table \ref{tab:experiment_sin_2}. This underestimation phenomenon for large $v$ becomes more evident as $s$ becomes larger, namely as the tail of Zipf's distribution becomes lighter and hence the fraction of symbols with low-frequency becomes smaller. For instance, we observe that for the bin $(128,256]$ the MAE for $s=2.5$ is larger than the MAE for $s=1.3$.

We also present an application of the CMS-NIGP to textual datasets, for which the distribution of words is typically a power-law distribution. See \cite{cla09} and references therein. Here, we consider the 20 Newsgroups dataset (\url{http://qwone.com/~jason/20Newsgroups/}) and the Enron dataset (\url{https://archive.ics.uci.edu/ml/machine-learning-databases/bag-of-words/}). The 20 Newsgroups dataset consists of $m=2.765.300$ tokens with $k=53.975$ distinct tokens, whereas the Enron dataset consists of $m=6412175$ tokens with $k=28102$ distinct tokens. Following experiments in \cite{cai2018bayesian}, we make use of a $2$-universal hash family, with the following hashing parameters: i) $J=12000$ and $N=2$; ii) $J=8000$ and $N=4$. By means the goodness of fit test proposed in \cite{cla09}, we found that the 20 Newsgroups and Enron datasets fit with a power-law distribution with exponent $s=2.3$ and $s=2.1$, respectively. Table \ref{tab:experiment_rea} reports the MAE of $\hat{f}_{v}^{\text{\tiny{(DP)}}}$ and $\hat{f}_{v}^{\text{\tiny{(NIGP)}}}$ applied to the 20 Newsgroups dataset and to the Enron dataset. Results of Table \ref{tab:experiment_rea} confirms the behaviour observed in Zipf' synthetic data. That is, $\hat{f}_{v}^{\text{\tiny{(NIGP)}}}$ outperforms $\hat{f}_{v}^{\text{\tiny{(DP)}}}$ for low-frequency tokens, whereas $\hat{f}_{v}^{\text{\tiny{(DP)}}}$ of \cite{cai2018bayesian} has a better performance than $\hat{f}_{v}^{\text{\tiny{(NIGP)}}}$ for high-frequency tokens. Table \ref{tab:experiment_rea} also contains a comparison with respect to $\hat{f}_{v}^{\text{\tiny{(CMM)}}}$, revealing that $\hat{f}_{v}^{\text{\tiny{(NIGP)}}}$ is competitive with $\hat{f}_{v}^{\text{\tiny{(CMM)}}}$ in the estimation of low-frequency tokens both the choices of hashing parameters.

\begin{table*}[!htb]
    \caption{Synthetic data: MAE for $\hat{f}_{v}^{\text{\tiny{(NIGP)}}}$ and $\hat{f}_{v}^{\text{\tiny{(DP)}}}$, case $J=320,N=2$}
    \label{tab:experiment_sin}
    \centering
    \resizebox{0.9\textwidth}{!}{\begin{tabular}{lrrrrrrrrrr}
        \toprule
        \multicolumn{1}{l}{} & \multicolumn{2}{c}{$\mathcal{Z}_{1.3}$} & \multicolumn{2}{c}{$\mathcal{Z}_{1.6}$} & \multicolumn{2}{c}{$\mathcal{Z}_{1.9}$} & \multicolumn{2}{c}{$\mathcal{Z}_{2.2}$} & \multicolumn{2}{c}{$\mathcal{Z}_{2.5}$} \\
        \cmidrule(r){2-3} \cmidrule(r){4-5} \cmidrule(r) {6-7} \cmidrule(r){8-9} \cmidrule(r){10-11}
        Bins $v$ & $\hat{f}_{v}^{\text{\tiny{(DP)}}}$ & $\hat{f}_{v}^{\text{\tiny{(NIGP)}}}$ & $\hat{f}_{v}^{\text{\tiny{(DP)}}}$ & $\hat{f}_{v}^{\text{\tiny{(NIGP)}}}$ & $\hat{f}_{v}^{\text{\tiny{(DP)}}}$ & $\hat{f}_{v}^{\text{\tiny{(NIGP)}}}$ & $\hat{f}_{v}^{\text{\tiny{(DP)}}}$ & $\hat{f}_{v}^{\text{\tiny{(NIGP)}}}$ & $\hat{f}_{v}^{\text{\tiny{(DP)}}}$ & $\hat{f}_{v}^{\text{\tiny{(NIGP)}}}$ \\[0.05cm]
        \midrule
            (0,1] & 1,057.61 & 231.31 &   626.85 & 134.75 &   306.70 &   65.71 &   51.38 &   12.91 & 32.43 &   7.16 \\
            (1,2] & 1,194.67 & 287.43 &   512.43 & 119.22 &   153.57 &   37.03 &  288.27 &   61.87 & 47.84 &   9.88 \\
            (2,4] & 1,105.16 & 262.18 &   472.59 &  95.78 & 2,406.00 &  353.73 &  133.31 &   26.90 & 53.97 &  10.09 \\
            (4,8] & 1,272.02 & 302.89 &   783.88 & 175.10 &   457.57 &   83.30 &  117.76 &   21.58 & 69.47 &  14.28 \\
           (8,16] & 1,231.63 & 257.08 &   716.52 & 136.66 &   377.99 &   66.44 &  411.21 &   77.39 & 80.43 &  20.15 \\
          (16,32] & 1,252.18 & 248.41 &   829.17 & 190.05 &   286.98 &   41.99 &  501.00 &   90.29 &  9.61 &  15.36 \\
          (32,64] & 1,309.14 & 284.12 &   780.70 & 139.52 &   413.95 &   67.30 &  216.84 &   48.00 &  9.89 &  28.90 \\
         (64,128] & 1,716.76 & 312.59 &   946.20 & 125.07 & 1,869.23 &  353.10 &   63.05 &   65.91 & 13.38 &  66.18 \\
        (128,256] & 1,102.96 &  97.91 & 1,720.49 & 273.50 &   199.87 &  110.32 &   45.98 &  130.94 & 17.03 & 125.75 \\[0.07cm]
        \bottomrule
    \end{tabular}}
\end{table*}

\begin{table*}[!htb]
    \caption{Synthetic data: MAE for $\hat{f}_{v}^{\text{\tiny{(NIGP)}}}$ and $\hat{f}_{v}^{\text{\tiny{(DP)}}}$, case $J=160,N=4$}
    \label{tab:experiment_sin_2}
    \centering
    \resizebox{0.9\textwidth}{!}{\begin{tabular}{lrrrrrrrrrr}
        \toprule
        \multicolumn{1}{l}{} & \multicolumn{2}{c}{$\mathcal{Z}_{1.3}$} & \multicolumn{2}{c}{$\mathcal{Z}_{1.6}$} & \multicolumn{2}{c}{$\mathcal{Z}_{1.9}$} & \multicolumn{2}{c}{$\mathcal{Z}_{2.2}$} & \multicolumn{2}{c}{$\mathcal{Z}_{2.5}$} \\
        \cmidrule(r){2-3} \cmidrule(r){4-5} \cmidrule(r) {6-7} \cmidrule(r){8-9} \cmidrule(r){10-11}
        Bins $v$ & $\hat{f}_{v}^{\text{\tiny{(DP)}}}$ & $\hat{f}_{v}^{\text{\tiny{(NIGP)}}}$ & $\hat{f}_{v}^{\text{\tiny{(DP)}}}$ & $\hat{f}_{v}^{\text{\tiny{(NIGP)}}}$ & $\hat{f}_{v}^{\text{\tiny{(DP)}}}$ & $\hat{f}_{v}^{\text{\tiny{(NIGP)}}}$ & $\hat{f}_{v}^{\text{\tiny{(DP)}}}$ & $\hat{f}_{v}^{\text{\tiny{(NIGP)}}}$ & $\hat{f}_{v}^{\text{\tiny{(DP)}}}$ & $\hat{f}_{v}^{\text{\tiny{(NIGP)}}}$ \\[0.05cm]
        \midrule
            (0,1] & 2,206.09 &   0.9 & 1,254.85 &   0.25 & 420.76 &   0.18 & 153.20 &   0.32 & 56.08 &   0.38 \\
            (1,2] & 2,333.06 &   0.5 & 1,326.71 &   0.70 & 549.12 &   0.82 & 180.71 &   1.24 & 47.48 &   1.45 \\
            (2,4] & 2,266.35 &   1.3 & 1,267.97 &   2.47 & 482.45 &   2.53 & 182.18 &   2.66 & 56.87 &   2.74 \\
            (4,8] & 2,229.22 &   4.6 & 1,371.27 &   4.67 & 538.91 &   5.28 & 250.32 &   5.96 & 50.30 &   5.42 \\
           (8,16] & 2,207.42 &  10.5 & 1,159.29 &  10.68 & 487.69 &  10.86 & 245.09 &  10.28 & 23.70 &  11.75 \\
          (16,32] & 2,279.80 &  20.7 & 1,211.41 &  19.21 & 529.77 &  22.08 & 293.68 &  21.57 & 24.41 &  23.37 \\
          (32,64] & 2,301.99 &  42.6 & 1,280.17 &  43.14 & 632.45 &  42.64 & 118.26 &  44.49 & 30.95 &  44.03 \\
         (64,128] & 2,241.57 &  92.2 & 1,112.41 &  94.43 & 419.42 &  95.19 & 177.61 &  95.10 & 28.78 &  93.34 \\
        (128,256] & 2,235.40 & 170.0 & 1,133.85 & 173.87 & 522.21 & 185.83 & 128.09 & 180.41 & 31.46 & 179.51 \\[0.07cm]
        \bottomrule
    \end{tabular}}
\end{table*}

\begin{table*}[!htb]
    \caption{Real data: MAE for $\hat{f}_{v}^{\text{\tiny{(NIGP)}}}$, $\hat{f}_{v}^{\text{\tiny{(DP)}}}$ and $\hat{f}_{v}^{\text{\tiny{(CMM)}}}$}
    \label{tab:experiment_rea}
    \centering
    \resizebox{\textwidth}{!}{\begin{tabular}{lrrrrrrrrrrrr}
        \toprule
        \multicolumn{1}{l}{} & \multicolumn{6}{c}{$J=12000$ and $N=2$} & \multicolumn{6}{c}{$J=8000$ and $N=4$}\\
        \cmidrule(r){2-7} \cmidrule(r){8-13}
        \multicolumn{1}{l}{} & \multicolumn{3}{c}{20 Newsgroups} & \multicolumn{3}{c}{Enron}& \multicolumn{3}{c}{20 Newsgroups} & \multicolumn{3}{c}{Enron} \\
        \cmidrule(r){2-4} \cmidrule(r){5-7}\cmidrule(r){8-10}\cmidrule(r){11-13}
         Bins $v$ & $\hat{f}_{v}^{\text{\tiny{(DP)}}}$ & $\hat{f}_{v}^{\text{\tiny{(NIGP)}}}$ & $\hat{f}_{v}^{\text{\tiny{(CMM)}}}$ & $\hat{f}_{v}^{\text{\tiny{(DP)}}}$ & $\hat{f}_{v}^{\text{\tiny{(NIGP)}}}$ & $\hat{f}_{v}^{\text{\tiny{(CMM)}}}$ & $\hat{f}_{v}^{\text{\tiny{(DP)}}}$ &  $\hat{f}_{v}^{\text{\tiny{(NIGP)}}}$ & $\hat{f}_{v}^{\text{\tiny{(CMM)}}}$ & $\hat{f}_{v}^{\text{\tiny{(DP)}}}$ & $\hat{f}_{v}^{\text{\tiny{(NIGP)}}}$ & $\hat{f}_{v}^{\text{\tiny{(CMM)}}}$\\[0.05cm]
        \midrule
                 (0,1] &  46.39 &  11.34 &    5.41 &  12.20 &   3.00 &    0.90 & 53.39 &   0.39 &    4.50 &  70.98 &   0.41 & 51.00\\
                 (1,2] &  16.60 &   3.53 &    2.16 &  13.80 &   3.06 &    2.00 & 30.49 &   1.40 &    2.00 &  47.38 &   1.47 & 27.20\\
                 (2,4] &  38.40 &   7.71 &    7.91 &  61.49 &  12.55 &    9.90 & 32.49 &   2.70 &    4.80 &  52.49 &   3.25 &  3.90\\
                 (4,8] &  59.39 &  10.40 &   35.70 &  88.39 &  17.36 &   17.32 & 38.69 &   5.97 &    6.23 &  53.08 &   6.17 & 10.50\\
                (8,16] &  54.29 &  11.34 &   45.40 &  23.40 &   4.58 &    9.52 & 25.29 &  11.97 &   13.50 &  56.98 &  11.28 & 22.20\\
               (16,32] &  17.80 &   9.85 &   20.99 &  55.09 &  11.58 &   21.00 & 24.99 &  21.25 &   21.60 &  89.98 &  19.82 & 20.60\\
               (32,64] &  40.79 &  25.65 &   58.86 & 128.48 &  39.46 &  134.47 & 39.69 &  42.81 &   39.22 & 108.37 &  47.07 & 61.38\\
              (64,128] &  25.99 &  57.95 &   91.59 & 131.08 &  54.42 &  110.27 & 22.09 &  91.06 &   86.32 &  55.67 &  87.32 & 66.50\\
             (128,256] &  13.59 & 126.07 &  186.92 &  50.68 & 119.04 &  140.43 & 25.79 & 205.58 &  183.96 &  80.76 & 178.23 & 90.20\\[0.07cm]
        \bottomrule
    \end{tabular}}
\end{table*}

\section{Discussion} \label{sec:dis}

Under the BNP approach to CMS of \cite{cai2018bayesian}, the restriction property of the DP is critical to compute the posterior distribution of a point query, given the hashed frequencies, whereas the finite-dimensional projective property of the DP is desirable for ease of estimating prior's parameters since it provides the likelihood function of the hashed frequencies. The NIGP prior is the sole discrete nonparametric prior with power-law behaviour that satisfies both the restriction property and the finite-dimensional projective property. This made the NIGP a somehow "forced" prior choice for our problem of extending the work of  \cite{cai2018bayesian} under power-law data streams. By relying on the restriction property and the finite-dimensional projective property of the NIGP, in this paper we have introduced the CMS-NIGP, which is a learning-augmented CMS under power-law data streams of token. The CMS-NIGP exploits BNP modeling to incorporate into the CMS, through the NIGP prior, a sensible power law behavior for the data stream. CMS-NIGP estimates of a point query are obtained as functionals, e.g. mean, mode, median, of the posterior distribution of the point query given the stored hashed frequencies. Applications to synthetic and real data have showed that the CMS-NIGP outperforms the CMS and the CMS-DP in the estimation of low-frequency tokens. The flaw of the CMS in the estimation of low-frequency token is quite well known, especially in the context of natural language processing \citep{Goy12,Goy09,Pit15}, and the CMS-NIGP is a new proposal to compensate for this flaw. In particular, the CMS-NIGPS results to be competitive with the CMM of \cite{Goy12} in the estimation of low-frequency token.

The NGGP  \citep{Jam02, Pru02, lijoi2007controlling} and the generalized negative Binomial process (GNBP) of \citep{Zou(16)} are examples of nonparametric priors with power-law behaviour, and we have considered them before turning to the NIGP. Both the NGGP and the GNBP have the restriction property which, however, leads to estimators of a point query that are more involved from a mathematical/computational perspective than estimators under the NIGP prior. See Appendix E. Both the NGGP and the GNBP do not have the finite-dimensional projective property. The lack of the finite-dimensional projective property makes impractical the use of the likelihood function of the hashed frequencies and, in addition, the complicated form of the predictive distributions induced by the NGGP the GNBP makes hard to apply likelihood-free methods to estimate prior's parameters. We are aware that the NIGP prior limits the flexibility of tuning the prior to the power-law degree of the data, in the sense that the NIGP is defined as a NGGP with $\sigma=1/2$. However, we believe that a NIGP prior is still a sensible choice of practical interest, especially in light of the fact that estimating $\sigma$ under the NGGP prior is a difficult task \citep{lijoi2007controlling}. Moreover, if one were forced to choose a specific value for $\sigma\in(0,1)$, without any information on the power-law degree of the data, $\sigma=1/2$ would arguably be a sensible and safe choice.

Many fruitful directions for future work remain open, especially with respect to the use of the BNP approach to develop learning-augmented CMSs that allows for adapting to the power-law degree of the data stream of tokens. Moreover, based on the promising empirical results of the BNP approach to CMS, we also encourage research to extend the BNP approach to other queries, e.g., range queries, inner product queries. This line of research would broaden the range of applications of the CMS, especially for data streams with power-law behaviour.

\section{Acknowledgment}

Stefano Favaro is also affiliated to IMATI-CNR ``Enrico Magenes" (Milan, Italy). Emanuele Dolera and Stefano Favaro received funding from the European Research Council (ERC) under the European Union's Horizon 2020 research and innovation programme under grant agreement No 817257. Emanuele Dolera and Stefano Favaro gratefully acknowledge the financial support from the Italian Ministry of Education, University and Research (MIUR), ``Dipartimenti di Eccellenza" grant agreement 2018-2022.

\bibliographystyle{apalike}
\bibliography{ref}

\end{document}


\onecolumn
\appendix

\section{The CMS}\label{app_cms} 
For any $m\geq1$ let $X_{1:m}=(X_{1},\ldots,X_{m})$ be a data stream of tokens taking values in a measurable space of symbols $\VV$. A point query over $X_{1:m}$ asks for the estimation of the frequency $f_{v}$ of a token of type $v\in\VV$ in $X_{1:m}$, i.e.  $f_v=\sum_{1\leq i\leq m} \ind_{X_{i}}(v)$. The goal of CMS of \citep{cormode2005summarizing,cor052} consists in estimating $f_{v}$ based on a compressed representation of $X_{1:m}$ by random hashing. In particular, let $J$ and $N$ be positive integers such that $[J]=\{1,\ldots,J\}$ and $[N]=\{1,\ldots,N\}$, and let $h_{1},\ldots,h_{N}$, with $h_{n}:\VV\rightarrow [J]$, be a collection of hash functions drawn uniformly at random from a pairwise independent hash family $\mathcal{H}$. That is, a random hash function $h\in\mathcal{H}$ has the property that for all $v_{1},v_{2}\in\mathcal{H}$ such that $v_{1}\neq v_{2}$, the probability that $v_{1}$ and $v_{2}$ hash to values $j_{1},j_{2}\in[J]$, respectively, is
\begin{displaymath}
\text{Pr}[h(v_{1})=j_{1},h(v_{2})=j_{2}]=\frac{1}{J^{2}}.
\end{displaymath}
Hashing $X_{1:m}$ through $h_{1},\ldots,h_{N}$ creates $N$ vectors of $J$ buckets $\{(C_{n,1},\ldots,C_{n,J})\}_{n\in[N]}$, with $C_{n,j}$ obtained by aggregating the frequencies for all $x$ where $h_{n}(x)=j$. Every $C_{n,j}$ is initialized at zero, and whenever a new token $X_i$ is observed we set $C_{n,h_n(X_i)} \leftarrow 1+ C_{n,h_n(X_i)}$ for every $n\in [N]$. After $m$ tokens, $C_{n,j}=\sum_{1\leq i\leq m} \ind_{h_n(X_i)}(j)$ and $f_v\le C_{n,j}$ for any $v \in \VV$. Under this setting, the CMS of \citep{cor052} estimates $f_{v}$ with the smallest hashed frequency among $\{C_{n,h_n(v)}\}_{n\in [N]}$, i.e.,
\begin{displaymath}
\hat{f}_{v}^{\text{\tiny{(CMS)}}}= \min_{n\in[N]}\{C_{n,h_{n}(v)}\}_{n\in[N]}.
\end{displaymath}
That is, $\hat{f}_{v}^{\text{\tiny{(CMS)}}}$ returns the count associated with the fewest collisions. This provides an upper bound on the true count. For an arbitrary data stream with $m$ tokens, the CMS satisfies the following guarantee.

\begin{theorem}
\citep{cor052} Let $J=\left \lceil{e/2}\right \rceil $ and let $N=\left \lceil{\log 1/\delta}\right \rceil$, with $\varepsilon>0$ and $\delta>0$. Then, the estimate $\hat{f}_{v}^{\text{\tiny{(CMS)}}}$ satisfies $\hat{f}_{v}^{\text{\tiny{(CMS)}}}\geq f_{v}$ and, with probability at least $1-\delta$, the estimate  $\hat{f}_{v}^{\text{\tiny{(CMS)}}}$ satisfies  $\hat{f}_{v}^{\text{\tiny{(CMS)}}}\leq f_{v}+\varepsilon m$.
\end{theorem}

\section{CRMs and hNCRMs}\label{app_ncrm}

Let $\VV$ be a measurable space endowed with its Borel $\sigma$-field $\mathcal{F}$. A CRM $\mu$ on $\VV$ is defined as a random measure such that for any $A_1,\ldots, A_k$ in $\mathcal{F}$, with $A_{i}\cap A_{j}=\emptyset$ for $i\neq j$, the random variables $\mu(A_1),\ldots,\mu(A_k)$ are mutually independent \citep{kingman2005poisson}. Any CRM $\mu$ with no fixed point of discontinuity and no deterministic drift is represented as $\mu=\sum_{j\geq1}\xi_{j}\delta_{v_{j}}$, where the $\xi_{j}$'s are positive random jumps and the $v_{j}$'s are $\VV$-valued random locations. Then, $\mu$ is characterized by the L\'evy--Khintchine representation
\begin{displaymath}
\EE\left[\exp\left\{-\int_{\VV}f(v)\mu(\text{d}v)\right\}\right]=\exp\left\{-\int_{\mathbb{R}^{+}\times\VV}[1-\text{e}^{-\xi f(v)}]\right\}\rho(\text{d}\xi,\text{d}v),
\end{displaymath}
where $f:\VV\rightarrow\mathbb{R}$ is a measurable function such that $\int|f|\text{d}\mu<+\infty$ and $\rho$ is a measure on $\RE^{+} \times \VV$ such that $\int_{B}\int_{\mathbb{R}^{+}}\min\{\xi,1\}\rho(\text{d}\xi,\text{d}v)<+\infty$ for any $B\in\mathcal{F}$. The measure $\rho$, referred to as L{\'e}vy intensity measure, characterizes $\mu$: it contains all the information on the distributions of jumps and locations of $\mu$.  For our purposes it will often be useful to separate the jump and location part of $\rho$ by writing it as
\begin{displaymath}
\gamma(\text{d}\xi,\text{d}v)=\rho(\text{d}\xi;v)\nu(\text{d}v),
\end{displaymath}
where $\nu$ denotes a measure on $(\mathcal{V},\mathcal{F})$ and $\rho$ denotes a transition kernel on $\mathcal{B}(\mathbb{R}^{+})\times\mathcal{V}$, with $\mathcal{B}(\mathbb{R}^{+})$ being the Borel $\sigma$-field of $\mathbb{R}^{+}$, i.e. $v\mapsto\rho(A;v)$ is $\mathcal{F}$-measurable for any $A\in\mathcal{B}(\mathbb{R}^{+})$ and $\rho(\cdot;v)$ is a measure on $(\mathbb{R}^{+},\mathcal{B}(\mathbb{R}^{+}))$ for any $v\in\mathcal{V}$. In particular, if $\rho(\cdot;v)=\rho(\cdot)$ for any $v$ then the jumps of $\mu$ is independent of their locations and $\gamma$ and $\mu$ are termed homogeneous. See \citep{kingman2005poisson} and references therein.

CRMs are closely connected to Poisson processes. Indeed $\mu$ can be represented as a linear functional of a Poisson process $\Pi$ on $\mathbb{R}^{+}\times\mathcal{V}$ with mean measure $\gamma$. To stated this precisely, $\Pi$ is a random subset of $\mathbb{R}^{+}\times\mathcal{V}$ and if $N(A)=\text{card}\{\Pi\cap A\}$ for any $A\subset\mathcal{B}(\mathbb{R}^{+})\otimes\mathcal{F}$ such that $\gamma(A)<+\infty$, then
\begin{displaymath}
\text{Pr}[N(A)=k]=\text{e}^{-\gamma(A)}\frac{(\gamma(A))^{k}}{k!}
\end{displaymath}
for $k\geq0$. Then, for any $A\in\mathcal{F}$
\begin{displaymath}
\mu(A)=\int_{A}\int_{\mathbb{R}^{+}}N(\text{d}v,\text{d}\xi)
\end{displaymath}
See \citep{kingman2005poisson} and references therein. An important property of CRMs is their almost sure discreteness \citep{kingman2005poisson}, which means that their realizations are discrete measures with probability $1$. This fact essentially entails discreteness of random probability measures obtained as transformations of CRMs, such as hNCRMs.

hNCRMs \citep{Jam02, Pru02,Reg03, pitman2006combinatorial,lij10} are defined in terms of a suitable normalization of CRMs. Let $\mu$ be a homogeneous CRM on $\mathcal{V}$ such that $0<\mu(\mathcal{V})<+\infty$ almost surely. Then, the random probability measure
\begin{equation}\label{eq_norm}
    P =\frac{\mu}{\mu(\VV)}
\end{equation}
is termed hNCRM. Because of the almost sure discreteness of $\mu$, the $P$ is discrete almost surely. That is, 
\begin{displaymath}
P=\sum_{j\geq1} p_j \delta_{v_j},
\end{displaymath}
where $p_{j}=\xi_{j}/\mu(\VV)$ for $j\geq1$ are random probabilities such that $p_{j}\in(0,1)$ for any $j\geq1$ and $\sum_{j\geq1}p_{j}=1$ almost surely. Both the conditions of finiteness and positiveness of $\mu(\mathcal{V})$ are clearly required for the normalization \eqref{eq_norm} to be well-defined, and it is natural to express these conditions in terms of the L{\'e}vy intensity measure $\gamma$ of the CRM $\mu$. It is enough to have $\rho=+\infty$ and $0<\mu(\mathcal{V})<+\infty$. In particular, the former is equivalent to requiring that $\mu$ has infinitely many jumps on any bounded set: in this case $\mu$ is also called an infinite activity process. The previous conditions can also be strengthened to necessary and sufficient conditions but we do not pursue this here. See \citep{kingman2005poisson}.

\section{NGGP priors, and proof of Proposition 1}\label{app_nggp}
Let $\VV$ be a measurable space endowed with its Borel $\sigma$-field $\mathcal{F}$. For any  $m\geq1$, let $X_{1:m}$ be a random sample of tokens from $P\sim\text{NGGP}(\alpha,\sigma,\nu)$. Because of the discreteness of $P$, the random sample $X_{1:m}$ induces a random partition of the set $\{1,\ldots,m\}$ into $K_{m}=k\leq m$ partition subsets, labelled by distinct symbols $\mathbf{v}=\{v_1,\ldots,v_{K_{m}}\}$ in $\VV$, with frequencies $\mathbf{N}_{n}=(N_{1},\ldots,N_{K_{m}})=(n_{1},\ldots,n_{k})$ such that $N_{i}>0$ and $\sum_{1\leq i\leq K_{m}}N_{i}=m$. Distributional properties of the random partition induced by $X_{1:m}$ induced by $X_{1:m}$ have been investigated in, e.g., \citep{Jam02}, \cite{pit03}, \citep{lijoi2007controlling}, \citep{de2013gibbs}  and \citep{Bac17}. In particular,
\begin{equation}\label{randpar_nggp}
\text{Pr}[K_{m}=k,\mathbf{N}_{m}=(n_{1},\ldots,n_{k})]=\frac{1}{k!}{m\choose n_{1},\ldots,n_{k}}V_{m,k}\prod_{i=1}^{k}(1-\sigma)_{(n_{i}-1)},
\end{equation}
where
\begin{equation}\label{weight_nggp}
V_{m,k}=\frac{(\alpha2^{\sigma-1})^{k}}{\Gamma(m)}\int_{0}^{+\infty}\frac{x^{m-1}}{(2^{-1}+x)^{m-k\sigma}}\exp\left\{-\frac{\alpha2^{\sigma-1}}{\sigma}[(2^{-1}+x)^{\sigma}-2^{-\sigma}]\right\}\de x.
\end{equation}
Now, let $\mathcal{P}_{m,k}=\{(n_{1},\ldots,n_{k})\text{ : }n_{i}\geq0\text{ and }\sum_{1\leq i\leq k}n_{i}=m\}$ denote the set of partitions of $m$ into $k\leq m$ blocks. Then, the distribution of $K_{m}$ follows my marginalizing \eqref{randpar_nggp} on the set $\mathcal{P}_{m,k}$, that is
\begin{align}\label{randk_nggp}
\notag\text{Pr}[K_{m}=k]&=\sum_{(n_{1},\ldots,n_{k})\in\mathcal{P}_{m,k}}\frac{1}{k!}{m\choose n_{1},\ldots,n_{k}}V_{m,k}\prod_{i=1}^{k}(1-\sigma)_{(n_{i}-1)}\\
&=\frac{V_{m,k}}{\sigma^{k}}C(m,k;\sigma),
\end{align}
where $C(m,k;\sigma)$ denotes the (central) generalized factorial coefficient \citep{charalambides2005combinatorial}, which is defined as $C(m,k;\sigma)=(k!)^{-1}\sum_{1\leq i\leq k}{k\choose i}(-1)^{i}(i\sigma)_{(m)}$, with the proviso $C(0,0;\sigma)=1$ and $C(m,0;\sigma)=0$ for any $m\geq1$. For any $1\leq r\leq m$, let $M_{r,m}\geq0$ denote the number of distinct symbols with frequency $r$ in $X_{1:m}$, i.e. $M_{r,m}=\sum_{1\leq i\leq K_{m}} \ind_{N_i}(r)$ such that $\sum_{1\leq r\leq m}M_{r,m}=K_{m}$ and $\sum_{1\leq r\leq m}rM_{r,m}=m$. Then, the distribution of $\mathbf{M}_{m}=(M_{1,m},\ldots,M_{m,m})$ follows directly form \eqref{randpar_nggp}, i.e.
\begin{equation}\label{eq:distributionapp}
\text{Pr}[\mathbf{M}_{m}=\mathbf{m}]=V_{m,k}m!\prod_{i=1}^{m}\left(\frac{(1-\sigma)_{(i-1)}}{i!}\right)^{m_{i}}\frac{1}{m_{i}!}\ind_{\mathcal{M}_{m,k}}(\mathbf{m}),
\end{equation}
where $\mathcal{M}_{m,k}=\{(m_{1},\ldots,m_{n})\text{ : }m_{i}\geq0,\text{ and }\sum_{1\leq i\leq m}m_{i}=k,\,\sum_{1\leq i\leq m}im_{i}=m\}$. The distribution \eqref{eq:distributionapp} is the referred to as the sampling formula of the random partition with distribution \eqref{randpar_nggp}.

For any $m\geq1$, let $X_{1:m}$ be a random sample from $P\sim\text{NGGP}(\alpha,\sigma,\nu)$ featuring $K_{m}=k$ partition subsets, labelled by distinct symbols $\mathbf{v}=\{v_1,\ldots,v_{K_{m}}\}$ in $\VV$, with frequencies $\mathbf{N}_{n}=(n_{1},\ldots,n_{k})$. The predictive distributions of $P$ provides the conditional distribution of $X_{m+1}$ given $X_{1:m}$. That is, for $A\in\mathcal{F}$
\begin{equation}\label{predictive_nggp}
\text{Pr}[X_{m+1}\in A\,|\,X_{1:m}]=\frac{V_{m+1,k+1}}{V_{m,k}}\nu(A)+\frac{V_{m+1,k}}{V_{m,k}}\sum_{i=1}^{k}(n_{i}-\sigma)\delta_{v_{i}}(A)
\end{equation}
for any $m\geq1$. We refer to \cite{Bac17} for a characterization of \eqref{predictive_nggp} in terms of a meaningful P\'olya like urn scheme. The predictive distributions \eqref{predictive_nggp} provides the fundamental ingredient of the proof of Proposition 1.

\textit{Proof of Proposition 1.} The proof follows from the predictive distributions \eqref{predictive_nggp} by setting $A=\mathbf{v}_{0}$ and $A=\mathbf{v}_{r}$.\qed

We conclude by showing that the distributional property of a random sample from $P\sim\text{DP}(\alpha,\nu)$ follows from the distributional property of a random sample from $P\sim\text{NGGP}(\alpha,\sigma,\nu)$ by letting $\sigma\rightarrow0$. For any $m\geq1$, let $X_{1:m}$ be a random sample from $P\sim\text{DP}(\alpha/2,\nu)$ featuring $K_{m}=k$ partition subsets, labelled by distinct symbols $\mathbf{v}=\{v_1,\ldots,v_{K_{m}}\}$ in $\VV$, with frequencies $\mathbf{N}_{n}=(n_{1},\ldots,n_{k})$. The distribution of the random partition induced by $X_{1:m}$ follows from \eqref{randpar_nggp} by letting $\sigma\rightarrow0$. Indeed,
\begin{align}\label{weight_dp}
\notag\lim_{\sigma\rightarrow0}V_{m,k}&=\lim_{\sigma\rightarrow+0}\frac{(\alpha2^{\sigma-1})^{k}}{\Gamma(m)}\int_{0}^{+\infty}\frac{x^{m-1}}{(2^{-1}+x)^{m-k\sigma}}\exp\left\{-\frac{\alpha2^{\sigma-1}}{\sigma}[(2^{-1}+x)^{\sigma}-2^{-\sigma}]\right\}\de x\\
&\notag=\frac{(\alpha/2)^{k}}{\Gamma(m)}2^{-\alpha/2}\int_{0}^{+\infty}\frac{x^{m-1}}{(2^{-1}+x)^{m+\alpha/2}}\ddr x\\
&=\frac{\left(\frac{\alpha}{2}\right)^{k}}{\left(\frac{\alpha}{2}\right)_{(m)}}.
\end{align}
Therefore, by combining the distribution \eqref{randpar_nggp} with \eqref{weight_dp}, and letting $\sigma\rightarrow0$, it follows directly the distribution of the random partition induced by a random sample $X_{1:m}$ from $P\sim\text{DP}(\alpha/2,\nu)$. That is, 
\begin{displaymath}
\text{Pr}[K_{m}=k,\mathbf{N}_{m}=(n_{1},\ldots,n_{k})]=\frac{1}{k!}{m\choose n_{1},\ldots,n_{k}}\frac{\left(\frac{\alpha}{2}\right)^{k}}{\left(\frac{\alpha}{2}\right)_{(m)}}\prod_{i=1}^{k}(n_{i}-1)!.
\end{displaymath}
The distribution of $K_{m}$ follows by combining the distribution \eqref{randk_nggp} with \eqref{weight_dp}, and from the fact that $\lim_{\sigma\rightarrow0}\sigma^{-k}C(m,k;\sigma)=|s(m,k)|$, where $|s(m,k)|$ denotes the signless Stirling number of the first type \citep{charalambides2005combinatorial}. That is, 
\begin{displaymath}
\text{Pr}[K_{m}=k]=\frac{\left(\frac{\alpha}{2}\right)^{k}}{\left(\frac{\alpha}{2}\right)_{(m)}}|s(m,k)|.
\end{displaymath}
In a similar manner, the distribution of $\mathbf{M}_{m}$ under the DP prior, which is referred to as Ewens sampling formula \cite{ewe72}, follows by combining the sampling formula \eqref{eq:distributionapp} with \eqref{weight_dp}, and letting $\sigma\rightarrow0$. 

Finally, the predictive distributions of $P\sim\text{DP}(\alpha,\nu)$. For any $m\geq1$, let $X_{1:m}$ be a random sample from $P\sim\text{DP}(\alpha/2,\nu)$ featuring $K_{m}=k$ partition subsets, labelled by distinct symbols $\mathbf{v}=\{v_1,\ldots,v_{K_{m}}\}$ in $\VV$, with frequencies $\mathbf{N}_{n}=(n_{1},\ldots,n_{k})$. The predictive distributions of $P$ follows by combining the predictive distributions \eqref{predictive_nggp} with \eqref{weight_dp}, and letting $\sigma\rightarrow0$. That is, for $A\in\mathcal{F}$
\begin{equation}\label{predictive_dp}
\text{Pr}[X_{m+1}\in A,|\,X_{1:m}]=\frac{\frac{\alpha}{2}}{\frac{\alpha}{2}+m}\nu(A)+\frac{1}{\frac{\alpha}{2}+m}\sum_{i=1}^{k}n_{i}\delta_{v_{i}}(A)
\end{equation}
for any $m\geq1$. The predictive distributions \eqref{predictive_dp} is at the basis of the CMS-DP proposed in \cite{cai2018bayesian}. In particular, Equation 4 in \cite{cai2018bayesian} follows from the predictive distributions \eqref{predictive_dp} by setting $A=\mathbf{v}_{0}$ and $A=\mathbf{v}_{r}$.

\section{The NIGP prior}\label{app_nig}
For $\sigma = 1/2$ the NGGP prior reduces to the NIGP prior \citep{Pru02,lijoi2005hierarchical}. Al alternative definition of the NIGP prior is given through its family of finite-dimensional distributions. This alternative definition relies on the IG distribution \citep{Ses93}. In particular, a random variable $W$ has IG distribution with shape parameter $a\geq0$ and scale parameter $b\geq0$ if it has the density function, with respect to the Lebesgue measure, given by
\begin{displaymath}
f_{W}(w;a,b)=\frac{a\text{e}^{ab}}{\sqrt{2\pi}}w^{-\frac{3}{2}}\exp\left\{-\frac{1}{2}\left(\frac{a^{2}}{w}+b^{2}w\right)\right\}\ind_{\mathbb{R}^{+}}(w).
\end{displaymath}
Let $(W_{1},\ldots,W_{k})$ be a collection of independent random variables such that $W_{i}$ is distributed according to the IG distribution with shape parameter $a_{i}$ and scale parameter $1$, for $i=1,\ldots,k$. The normalized IG distribution with parameter $(a_{1},\ldots,a_{k})$ is the distribution of the following random variable 
\begin{displaymath}
(P_{1},\ldots,P_{k})=\left(\frac{W_{1}}{\sum_{i=1}^{k}W_{i}},\ldots,\frac{W_{k}}{\sum_{i=1}^{k}W_{i}}\right).
\end{displaymath}
The distribution of the random variable $(P_{1},\ldots,P_{k-1})$ is absolutely continuous with respect to the Lebesgue measure on $R^{k-1}$, and its density function on the $(k-1)$-dimensional simplex coincides with
\begin{align}\label{norm_ig}
&f_{(P_{1},\ldots,P_{k-1})}(p_{1},\ldots,p_{k-1};a_{1},\ldots,a_{k})\\
&\notag\quad=\left(\prod_{i=1}^{k}\frac{a_{i}\text{e}^{a_{i}}}{\sqrt{2\pi}}\right)\prod_{i=1}^{k-1}p_{i}^{-3/2}\left(1-\sum_{i=1}^{k-1}p_{i}\right)^{-3/2}\\
&\notag\quad\quad\times2\left(\sum_{i=1}^{k-1}\frac{a_{i}^{2}}{p_{i}}+\frac{a_{k}^{2}}{1-\sum_{i=1}^{k-1}p_{i}}\right)^{-k/4}K_{-k/2}\left(\sqrt{\sum_{i=1}^{k-1}\frac{a_{i}^{2}}{p_{i}}+\frac{a_{k}^{2}}{1-\sum_{i=1}^{k-1}p_{i}}}\right),
\end{align}
where $K_{-k/2}$ denotes the modified Bessel function of the second type, or Macdonald function, with parameter $-k/2$. If the random variable $(P_{1},\ldots,P_{k})$ is distributed according to a normalized IG distribution with parameter $(a_{1},\ldots,a_{k})$, and if $m_{1}<m_{2}<\cdots<m_{r}<k$ are positive integers, then
\begin{displaymath}
\left(\sum_{i=1}^{m_{1}}P_{i},\sum_{i=m_{1}+1}^{m_{2}}P_{i},\ldots,\sum_{i=m_{r-1}+1}^{k}W_{i}\right)
\end{displaymath}
is a random variable distributed as a normalized inverse Gaussian distribution with parameter $(\sum_{1\leq i\leq m_{1}}a_{i},\sum_{m_{1}+1\leq i \leq m_{2} }a_{i},\ldots,\sum_{m_{r-1}+1\leq i\leq k}^{k}W_{i})$. This projective property of the normalized inverse Gaussian distribution follows from the additive property of the inverse Gaussian distribution \citep{Ses93}.

To define the NIGP prior through its family of finite-dimensional distributions, let $\VV$ be a measurable space endowed with its Borel $\sigma$-field $\mathcal{F}$. Let $\mathcal{P}=\{Q_{B_{1},\ldots,B_{k}}\text{ : }B_{1},\ldots,B_{k}\in\mathcal{F}\text{ for } k\geq1\}$ be a family of probability distributions, and let $\tilde{\nu}=\alpha\nu$ be a diffuse (base) measure on $\VV$ with $\tilde{\nu}(\VV)=\alpha$. If $\{B_{1},\ldots,B_{k}\}$ denotes a measurable $k$-partition of $\VV$ and $\Delta_{k-1}$ is the $(k-1)$-dimensional simplex, then set
\begin{displaymath} 
Q_{B_{1},\ldots,B_{k}}(C)=\int_{C\cap\Delta_{k-1}}f_{(P_{1},\ldots,P_{k-1})}(p_{1},\ldots,p_{k-1};a_{1},\ldots,a_{k})\ddr p_{1}\cdots \ddr p_{k-1}
\end{displaymath}
for any $C$ in the Borel $\sigma$-field of $\mathbb{R}^{k}$, where $f_{(P_{1},\ldots,P_{k-1})}$ is the normalized IG distribution with  density function \eqref{norm_ig} with $a_{i}=\tilde{\nu}(B_{i})$, for $i=1,\ldots,k$. According to Proposition 3.9.2 of \cite{Reg01}, the NIGP is the unique random probability measure admitting $\mathcal{P}$ as its family of finite-dimensional distributions.

The projective property of $P\sim\text{NIGP}(\alpha,\nu)$ follows directly from: i) the definition of $P$ through its family of finite-dimensional distributions; ii) the projective property of the normalized IG distribution. In particular, for any finite family of sets $\{A_{1},\ldots,A_{k}\}$ in $\mathcal{F}$, let $\{B_{1},\ldots,B_{h}\}$ be a measurable $h$-partition of $\VV$ such that it is finer then the partition generated by the family of sets $\{A_{1},\ldots,A_{k}\}$. Then,
\begin{displaymath}
Q_{A_{1},\ldots,A_{k}}(C)=Q_{B_{1},\ldots,B_{h}}(C^{\prime})
\end{displaymath}
for any $C$ in the Borel $\sigma$-field of $\mathbb{R}^{k}$, with $C^{\prime}=\{(x_{1},\ldots,x_{h})\in[0,1]^{h}\text{ : }(\sum_{i}x_{i},\ldots,\sum_{i}x_{i})\in C\}$. See \citep{lijoi2005hierarchical}.

\section{Proof of Proposition 2, and proof of Theorem 3}\label{app_proof_teo}

To prove Proposition 2, we start with the following lemma under the assumption that $X_{1:m}$ is a random sample from $P\sim\text{NGGP}(\alpha,\sigma,\nu)$. The proof of Proposition 2 then follows by setting $\sigma=1/2$. Let 
\begin{displaymath}
p_{f_{v}}(\ell;m,\alpha,\sigma)=\sum_{\mathbf{m}\in\mathcal{M}_{k,m}}\text{Pr}[X_{m+1} \in \mathbf{v}_{\ell} \mid \mathbf{M}_{m}=\mathbf{m}]\text{Pr}[\mathbf{M}_{m}=\mathbf{m}],\quad\ell=0,1,\ldots,m,
\end{displaymath}
where the predictive distributions $\text{Pr}[X_{m+1} \in \mathbf{v}_{\ell} \mid \mathbf{M}_{m}=\mathbf{m}]$ are displayed in Equation 5, and the distribution $\text{Pr}[\mathbf{M}_{m}=\mathbf{m}]$ is displayed in Equation 4. For $\sigma\in(0,1)$, let $f_{\sigma}$ denote the density function of the positive $\sigma$-stable random variable $X_{\sigma}$, i.e. $\mathbb{E}[\exp\{-tX_{\sigma}\}]=\exp\{-t^{\sigma}\}$ for any $t>0$.

\begin{lemma}\label{lem_ngg}
For any $m\geq1$, let $X_{1:m}$ be a random sample from $P\sim\text{NGGP}(\alpha,\sigma,\nu)$. Then, for $\ell=0,1,\ldots,m$
\begin{align}\label{probab_nggp}
p_{f_{v}}(\ell;m,\alpha,\sigma)
\begin{cases} 
\frac{\sigma(\ell-\sigma){m\choose \ell}(1-\sigma)_{(\ell-1)}}{\Gamma(1-\sigma+\ell)}\\
\quad\times\int_{0}^{+\infty}\int_{0}^{1}\frac{1}{h^{\sigma}}f_{\sigma}(hp)\text{e}^{-h\left(\frac{\alpha2^{-1}}{\sigma}\right)^{\frac{1}{\sigma}}+\frac{\alpha2^{-1}}{\sigma}}p^{m-\ell}(1-p)^{1-\sigma+\ell-1}\ddr p\ddr h&\quad\ell<m\\[0.4cm] 
\frac{\alpha2^{m}(1-\sigma)_{(m)}}{\Gamma(m+1)}\\
\quad\times\int_{0}^{+\infty}\frac{x^{m}}{(1+2x)^{m+1-\sigma}}\exp\left\{-\frac{\alpha2^{\sigma-1}}{\sigma}[(2^{-1}+x)^{\sigma}-2^{-\sigma}]\right\}\de x&\quad \ell=m.
\end{cases}
\end{align}
\end{lemma}

\begin{proof}
We start by considering the case $\ell=0$. The probability $p_{f_{v}}(0;m,\alpha,\sigma)$ follows by combining Proposition 1 with the distribution of $K_{m}$ displayed in \eqref{randk_nggp}. Indeed, we can write the following expression
\begin{align}\label{proof1_1}
p_{f_{v}}(0;m,\alpha,\sigma)&=\sum_{\mathbf{m}\in\mathcal{M}_{m,k}}\text{Pr}[X_{m+1} \in \mathbf{v}_{0} \mid \mathbf{M}_{m}=\mathbf{m}]\text{Pr}[\mathbf{M}_{m}=\mathbf{m}]\\
&\notag=\sum_{\mathbf{m}\in\mathcal{M}_{m,k}}\frac{V_{m+1,k+1}}{V_{m,k}}\text{Pr}[\mathbf{M}_{m}=\mathbf{m}]\\
&=\sum_{k=1}^{m}\frac{V_{m+1,k+1}}{\sigma^{k}}C(m,k;\sigma).
\end{align}
Then, the expression of $p_{f_{v}}(0;m,\alpha,\sigma)$ in \eqref{probab_nggp} follows by combining \eqref{proof1_1} with $V_{m+1,k+1}$ displayed in \eqref{weight_nggp}, i.e.,
\begin{align*}
p_{f_{v}}(0;m,\alpha,\sigma)&=\frac{(\alpha2^{\sigma-1})}{\Gamma(m+1)}\int_{0}^{+\infty}\frac{u^{m}}{(2^{-1}+u)^{m+1-\sigma}}\exp\left\{-\frac{\alpha2^{\sigma-1}}{\sigma}[(2^{-1}+u)^{\sigma}-2^{-\sigma}]\right\}\\
&\quad\times\sum_{k=1}^{m}\left(\frac{\alpha2^{\sigma-1}}{\sigma(2^{-1}+u)^{-\sigma}}\right)^{k}C(m,k;\sigma)\ddr u\\
&\text{[Equation 13 of \cite{fav15}]}\\
&=\frac{(\alpha2^{\sigma-1})}{\Gamma(m+1)}\int_{0}^{+\infty}\frac{u^{m}}{(2^{-1}+u)^{m+1-\sigma}}\exp\left\{-\frac{\alpha2^{\sigma-1}}{\sigma}[(2^{-1}+u)^{\sigma}-2^{-\sigma}]\right\}\\
&\quad\times\exp\left\{\frac{\alpha2^{\sigma-1}}{\sigma(2^{-1}+u)^{-\sigma}}\right\}\left(\frac{\alpha2^{\sigma-1}}{\sigma(2^{-1}+u)^{-\sigma}}\right)^{m/\sigma}\int_{0}^{+\infty}x^{m}\exp\left\{-x\left(\frac{\alpha2^{\sigma-1}}{\sigma(2^{-1}+u)^{-\sigma}}\right)^{1/\sigma}\right\}f_{\sigma}(x)\ddr x \ddr u\\
&[\text{Identity }(2^{-1}+u)^{-1+\sigma}=\frac{1}{\Gamma(1-\sigma)}\int_{0}^{+\infty}y^{1-\sigma-1}\exp\left\{-y(2^{-1}+u)\right\}\ddr y]\\
&=\frac{(\alpha2^{\sigma-1})^{1+m/\sigma}}{\sigma^{m/\sigma}\Gamma(m+1)}\\
&\quad\times\int_{0}^{+\infty}u^{m}\left(\frac{1}{\Gamma(1-\sigma)}\int_{0}^{+\infty}y^{1-\sigma-1}\exp\left\{-y(2^{-1}+u)\right\}\ddr y\right)\\
&\quad\quad\times\left(\int_{0}^{+\infty}x^{m}\exp\left\{-x\left(\frac{\alpha2^{\sigma-1}}{\sigma}\right)^{\frac{1}{\sigma}}u\right\}\exp\left\{-x\left(\frac{\alpha2^{-1}}{\sigma}\right)^{\frac{1}{\sigma}}+\frac{\alpha2^{-1}}{\sigma}\right\} f_{\sigma}(x)\ddr x\right)\ddr u\\
&=\frac{(\alpha2^{\sigma-1})^{1+m/\sigma}}{\sigma^{m/\sigma}\Gamma(1-\sigma)}\\
&\quad\times\int_{0}^{+\infty}x^{m}f_{\sigma}(x)\exp\left\{-x\left(\frac{\alpha2^{-1}}{\sigma}\right)^{\frac{1}{\sigma}}+\frac{\alpha2^{-1}}{\sigma}\right\}\\
&\quad\quad\times\int_{0}^{+\infty}y^{1-\sigma-1}\exp\left\{-y2^{-1}\right\}\left[x\left(\frac{\alpha2^{\sigma-1}}{\sigma}\right)^{\frac{1}{\sigma}}+y\right]^{-n-1}\ddr y\ddr x\\
&[\text{Change of variable }p=\frac{y}{x\left(\frac{\alpha2^{\sigma-1}}{\sigma}\right)^{\frac{1}{\sigma}}+y}]\\
&=\frac{(\alpha2^{\sigma-1})^{1+m/\sigma}}{\sigma^{m/\sigma}\Gamma(1-\sigma)}\\
&\quad\times\int_{0}^{+\infty}x^{m}f_{\sigma}(x)\exp\left\{-x\left(\frac{\alpha2^{-1}}{\sigma}\right)^{\frac{1}{\sigma}}+\frac{\alpha2^{-1}}{\sigma}\right\}\\
&\quad\quad\quad\times\int_{0}^{1}\left(\frac{\left(\frac{\alpha2^{\sigma-1}}{\sigma}\right)^{\frac{1}{\sigma}}xp}{1-p}\right)^{1-\sigma-1}\frac{\left(\frac{\alpha2^{\sigma-1}}{\sigma}\right)^{\frac{1}{\sigma}}x}{(1-p)^{2}}\\
&\quad\quad\quad\quad\times\exp\left\{-\frac{\left(\frac{\alpha2^{\sigma-1}}{\sigma}\right)^{\frac{1}{\sigma}}xp}{1-p}2^{-1}\right\}\left[x\left(\frac{\alpha2^{\sigma-1}}{\sigma}\right)^{\frac{1}{\sigma}}+\frac{\left(\frac{\alpha2^{\sigma-1}}{\sigma}\right)^{\frac{1}{\sigma}}xp}{1-p}\right]^{-m-1}\ddr p\ddr x\\
&[\text{Change of variable }h=x/(1-p)]\\
&=\frac{(\alpha2^{\sigma-1})^{1+m/\sigma}}{\sigma^{m/\sigma}\Gamma(1-\sigma)}\\
&\quad\times\int_{0}^{+\infty}\int_{0}^{1}(h(1-p))^{m}f_{\sigma}(h(1-p))\exp\left\{-h(1-p)\left(\frac{\alpha2^{-1}}{\sigma}\right)^{\frac{1}{\sigma}}+\frac{\alpha2^{-1}}{\sigma}\right\}\\
&\quad\quad\quad\times\left(\left(\frac{\alpha2^{\sigma-1}}{\sigma}\right)^{\frac{1}{\sigma}}hp\right)^{1-\sigma-1}\left(\frac{\alpha2^{\sigma-1}}{\sigma}\right)^{\frac{1}{\sigma}}h\\
&\quad\quad\quad\quad\times\exp\left\{-\left(\frac{\alpha2^{\sigma-1}}{\sigma}\right)^{\frac{1}{\sigma}}hp2^{-1}\right\}\left[h(1-p)\left(\frac{\alpha2^{\sigma-1}}{\sigma}\right)^{\frac{1}{\sigma}}+\left(\frac{\alpha2^{\sigma-1}}{\sigma}\right)^{\frac{1}{\sigma}}hp\right]^{-m-1}\ddr p\ddr h\\
&=\frac{(\alpha2^{\sigma-1})^{1+m/\sigma}}{\sigma^{m/\sigma}\Gamma(1-\sigma)}\left(\left(\frac{\alpha2^{\sigma-1}}{\sigma}\right)^{\frac{1}{\sigma}}\right)^{-\sigma-m}\\
&\quad\times\int_{0}^{+\infty}\int_{0}^{1}(h(1-p))^{n}f_{\sigma}(h(1-p))\exp\left\{-h\left(\frac{\alpha2^{-1}}{\sigma}\right)^{\frac{1}{\sigma}}+\frac{\alpha2^{-1}}{\sigma}\right\}p^{1-\sigma-1}h^{-m-\sigma}\ddr p \ddr h\\
&=\frac{\sigma}{\Gamma(1-\sigma)}\int_{0}^{+\infty}\int_{0}^{1}f_{\sigma}(hp)\exp\left\{-h\left(\frac{\alpha2^{-1}}{\sigma}\right)^{\frac{1}{\sigma}}+\frac{\alpha2^{-1}}{\sigma}\right\}h^{-\sigma}p^{m}(1-p)^{1-\sigma-1}\ddr p \ddr h.
\end{align*}
This complete the case $\ell=0$. Now, we consider $\ell>0$. The probability $p_{f_{v}}(\ell;m,\alpha,\sigma)$ follows by combining Proposition 1 with the distribution of $(K_{m},\mathbf{N}_{n})$ displayed in \eqref{randpar_nggp}. In particular, we can write
\begin{align}\label{proof1_2}
\notag p_{f_{v}}(\ell;m,\alpha,\sigma)&=\sum_{\mathbf{m}\in\mathcal{M}_{m,k}}\text{Pr}[X_{m+1} \in \mathbf{v}_{\ell} \mid\mathbf{M}_{m}=\mathbf{m}]\text{Pr}[\mathbf{M}_{m}=\mathbf{m}]\\
&\notag=\sum_{\mathbf{m}\in\mathcal{M}_{m,k}}\frac{V_{m+1,k}}{V_{m,k}}(\ell-\sigma)m_{l}\text{Pr}[\mathbf{M}_{m}=\mathbf{m}]\\
&\notag=(\ell-\sigma)\sum_{k=1}^{m}\sum_{(n_{1},\ldots,n_{k})\in \mathcal{P}_{m,k}}\frac{1}{k!}{m\choose n_{1},\ldots,n_{k}}V_{m,k}\prod_{i=1}^{k}(1-\sigma)_{(n_{i}-1)}\frac{V_{m+1,k}}{V_{m,k}}\sum_{j=1}^{k}\ind_{n_{j}}(\ell)\\
&\notag\quad=(\ell-\sigma)\sum_{k=1}^{m}\frac{V_{m+1,k}}{V_{m,k}}\sum_{j=1}^{k}\text{Pr}[K_{m}=k,N_{j}=\ell]\\
&\notag\quad=(\ell-\sigma)\sum_{k=1}^{m}\frac{V_{m+1,k}}{V_{m,k}}\sum_{j=1}^{k}\frac{V_{m,k}}{k}{n\choose \ell}(1-\sigma)_{(\ell-1)}\frac{C(m-\ell,k-1;\sigma)}{\sigma^{k-1}}\\
&\quad=(\ell-\sigma){m\choose \ell}(1-\sigma)_{(\ell-1)}\sum_{k=1}^{m}V_{m+1,k}\frac{C(m-\ell,k-1;\sigma)}{\sigma^{k-1}}.
\end{align}
Then, the expression of $p_{f_{v}}(\ell;m,\alpha,\sigma)$ in \eqref{probab_nggp} follows by combining \eqref{proof1_2} with $V_{m+1,k}$ displayed in \eqref{weight_nggp}, i.e., 
\begin{align*}
p_{f_{v}}(\ell;m,\alpha,\sigma)&=(\ell-\sigma){m\choose \ell}(1-\sigma)_{(\ell-1)}\sum_{k=1}^{m}V_{m+1,k}\frac{C(m-\ell,k-1;\sigma)}{\sigma^{k-1}}\\
&=(\ell-\sigma){m\choose \ell}(1-\sigma)_{(\ell-1)}\\
&\quad\times\frac{\sigma}{\Gamma(m+1)}\int_{0}^{+\infty}u^{m}\exp\left\{-\frac{\alpha2^{\sigma-1}}{\sigma}[(2^{-1}+u)^{\sigma}-2^{-\sigma}]\right\}(2^{-1}+u)^{-m-1}\ddr u\\
&\quad\quad\quad\times\sum_{k=1}^{m}C(m-\ell,k-1;\sigma)\left(\frac{\alpha2^{\sigma-1}}{\sigma(2^{-1}+u)^{-\sigma}}\right)^{k}.
\end{align*}
If $\ell=m$, then
\begin{align*}
p_{f_{v}}(m;m,\alpha,\sigma)&=(1-\sigma)_{(m)}\sum_{k=1}^{m}V_{m+1,k}\frac{C(0,k-1;\sigma)}{\sigma^{k-1}}\\
&=(1-\sigma)_{(m)}V_{m+1,1}\\
&=(1-\sigma)_{(m)}\frac{\alpha2^{\sigma-1}}{\Gamma(m+1)}\int_{0}^{+\infty}\frac{x^{m}}{(2^{-1}+x)^{m+1-\sigma}}\exp\left\{-\frac{\alpha2^{\sigma-1}}{\sigma}[(2^{-1}+x)^{\sigma}-2^{-\sigma}]\right\}\de x.
\end{align*}
If $\ell<m$, then
\begin{align*}
p_{f_{v}}(\ell;m,\alpha,\sigma)&=(\ell-\sigma){m\choose \ell}(1-\sigma)_{(\ell-1)}\sum_{k=1}^{m}V_{m+1,k}\frac{C(m-\ell,k-1;\sigma)}{\sigma^{k-1}}\\
&=(\ell-\sigma){m\choose \ell}(1-\sigma)_{(\ell-1)}\\
&\quad\times\frac{\alpha2^{\sigma-1}}{\Gamma(m+1)}\int_{0}^{+\infty}u^{m}\exp\left\{-\frac{\alpha2^{\sigma-1}}{\sigma}[(2^{-1}+u)^{\sigma}-2^{-\sigma}]\right\}(2^{-1}+u)^{-m-1+\sigma}\ddr u\\
&\quad\quad\times\sum_{k=1}^{m-\ell}C(m-\ell,k;\sigma)\left(\frac{\alpha2^{\sigma-1}}{\sigma(2^{-1}+u)^{-\sigma}}\right)^{k}\\
&\text{[Equation 13 of \cite{fav15}]}\\
&=(\ell-\sigma){m\choose \ell}(1-\sigma)_{(\ell-1)}\\
&\quad\times\frac{\alpha2^{\sigma-1}}{\Gamma(m+1)}\int_{0}^{+\infty}u^{m}\exp\left\{-\frac{\alpha2^{\sigma-1}}{\sigma}[(2^{-1}+u)^{\sigma}-2^{-\sigma}]\right\}(2^{-1}+u)^{-m-1+\sigma}\\
&\quad\quad\times\exp\left\{\frac{\alpha2^{\sigma-1}}{\sigma(2^{-1}+u)^{-\sigma}}\right\}\left(\frac{\alpha2^{\sigma-1}}{\sigma(2^{-1}+u)^{-\sigma}}\right)^{\frac{m-\ell}{\sigma}}\\
&\quad\quad\quad\times\int_{0}^{+\infty}x^{m-\ell}\exp\left\{-x\left(\frac{\alpha2^{\sigma-1}}{\sigma(2^{-1}+u)^{-\sigma}}\right)^{\frac{1}{\sigma}}\right\}f_{\sigma}(x)\ddr x\ddr u\\
&[\text{Identity }(2^{-1}+u)^{-1+\sigma}=\frac{1}{\Gamma(1-\sigma+\ell)}\int_{0}^{+\infty}y^{1-\sigma+\ell-1}\exp\left\{-y(2^{-1}+u)\right\}\ddr y]\\
&=(\ell-\sigma){m\choose \ell}(1-\sigma)_{(\ell-1)}\\
&\quad\times\frac{(\alpha2^{\sigma-1})(\alpha2^{\sigma-1})^{\frac{m-\ell}{\sigma}}}{\sigma^{\frac{m-\ell}{\sigma}}\Gamma(m+1)}\int_{0}^{+\infty}u^{m}\left(\frac{1}{\Gamma(1-\sigma+\ell)}\int_{0}^{+\infty}y^{1-\sigma+\ell-1}\exp\{-y(2^{-1}+u)\}\ddr y\right)\\
&\quad\quad\times\int_{0}^{+\infty}x^{m-\ell}\exp\left\{-xu\left(\frac{\alpha2^{\sigma-1}}{\sigma}\right)^{\frac{1}{\sigma}}\right\}\exp\left\{-x\left(\frac{\alpha2^{-1}}{\sigma}\right)^{\frac{1}{\sigma}}+\frac{\alpha2^{-1}}{\sigma}\right\}f_{\sigma}(x)\ddr x\ddr u\\
&=(\ell-\sigma){m\choose \ell}(1-\sigma)_{(\ell-1)}\\
&\times\frac{(\alpha2^{\sigma-1})^{1+\frac{m-\ell}{\sigma}}}{\sigma^{\frac{m-\ell}{\sigma}}\Gamma(1-\sigma+\ell)}\int_{0}^{+\infty}x^{m-\ell}f_{\sigma}(x)\exp\left\{-x\left(\frac{\alpha2^{-1}}{\sigma}\right)^{\frac{1}{\sigma}}+\frac{\alpha2^{-1}}{\sigma}\right\}\\
&\quad\quad\quad\times\int_{0}^{+\infty}y^{1-\sigma+\ell-1}\exp\{-y2^{-1}\}\left[x\left(\frac{\alpha2^{\sigma-1}}{\sigma}\right)^{\frac{1}{\sigma}}+y\right]^{-m-1}\ddr y\ddr x\\
&[\text{Change of variable }p=\frac{y}{x\left(\frac{\alpha2^{\sigma-1}}{\sigma}\right)^{\frac{1}{\sigma}}+y}]\\
&=(\ell-\sigma){m\choose \ell}(1-\sigma)_{(\ell-1)}\\
&\quad\times\frac{(\alpha2^{\sigma-1})^{1+\frac{m-\ell}{\sigma}}}{\sigma^{\frac{m-\ell}{\sigma}}\Gamma(1-\sigma+\ell)}\int_{0}^{+\infty}x^{m-\ell}f_{\sigma}(x)\exp\left\{-x\left(\frac{\alpha2^{-1}}{\sigma}\right)^{\frac{1}{\sigma}}+\frac{\alpha2^{-1}}{\sigma}\right\}\\
&\quad\quad\quad\times\int_{0}^{1}\left(\frac{\left(\frac{\alpha2^{\sigma-1}}{\sigma}\right)^{\frac{1}{\sigma}}xp}{1-p}\right)^{1-\sigma+\ell-1}\frac{\left(\frac{\alpha2^{\sigma-1}}{\sigma}\right)^{\frac{1}{\sigma}}x}{(1-p)^{2}}\\
&\quad\quad\quad\quad\times\exp\left\{-\frac{\left(\frac{\alpha2^{\sigma-1}}{\sigma}\right)^{\frac{1}{\sigma}}xp}{1-p}2^{-1}\right\}\left[x\left(\frac{\alpha2^{\sigma-1}}{\sigma}\right)^{\frac{1}{\sigma}}+\frac{\left(\frac{\alpha2^{\sigma-1}}{\sigma}\right)^{\frac{1}{\sigma}}xp}{1-p}\right]^{-m-1}\ddr p\ddr x\\
&[\text{Change of variable }h=x/(1-p)]\\
&=(\ell-\sigma){m\choose \ell}(1-\sigma)_{(\ell-1)}\\
&\quad\times\frac{(\alpha2^{\sigma-1})^{1+\frac{m-\ell}{\sigma}}}{\sigma^{\frac{m-\ell}{\sigma}}\Gamma(1-\sigma+\ell)}\int_{0}^{+\infty}\int_{0}^{1}(h(1-p))^{m-\ell}f_{\sigma}(h(1-p))\exp\left\{-h(1-p)\left(\frac{\alpha2^{-1}}{\sigma}\right)^{\frac{1}{\sigma}}+\frac{\alpha2^{-1}}{\sigma}\right\}\\
&\quad\quad\quad\times\left(\left(\frac{\alpha2^{\sigma-1}}{\sigma}\right)^{\frac{1}{\sigma}}hp\right)^{1-\sigma+\ell-1}\left(\frac{\alpha2^{\sigma-1}}{\sigma}\right)^{\frac{1}{\sigma}}h\\
&\quad\quad\quad\quad\times\exp\left\{-\left(\frac{\alpha2^{\sigma-1}}{\sigma}\right)^{\frac{1}{\sigma}}hp2^{-1}\right\}\left[h(1-p)\left(\frac{\alpha2^{\sigma-1}}{\sigma}\right)^{\frac{1}{\sigma}}+\left(\frac{\alpha2^{\sigma-1}}{\sigma}\right)^{\frac{1}{\sigma}}hp\right]^{-m-1}\ddr p\ddr h\\
&=(\ell-\sigma){m\choose \ell}(1-\sigma)_{(\ell-1)}\\
&\quad\times\frac{(\alpha2^{\sigma-1})^{1+\frac{m-\ell}{\sigma}}}{\sigma^{\frac{m-\ell}{\sigma}}\Gamma(1-\sigma+\ell)}\left(\left(\frac{\alpha2^{\sigma-1}}{\sigma}\right)^{\frac{1}{\sigma}}\right)^{-m-\sigma+\ell}\\
&\quad\quad\times\int_{0}^{+\infty}\int_{0}^{1}(h(1-p))^{m-\ell}f_{\sigma}(h(1-p))\exp\left\{-h\left(\frac{\alpha2^{-1}}{\sigma}\right)^{\frac{1}{\sigma}}+\frac{\alpha2^{-1}}{\sigma}\right\}p^{1-\sigma+\ell-1}h^{-m-\sigma+\ell}\ddr p\ddr h\\
&=(\ell-\sigma){m\choose \ell}(1-\sigma)_{(\ell-1)}\\
&\quad\times\frac{\sigma}{\Gamma(1-\sigma+\ell)}\int_{0}^{+\infty}\int_{0}^{1}f_{\sigma}(hp)\exp\left\{-h\left(\frac{\alpha2^{-1}}{\sigma}\right)^{\frac{1}{\sigma}}+\frac{\alpha2^{-1}}{\sigma}\right\}h^{-\sigma}p^{m-\ell}(1-p)^{1-\sigma+\ell-1}\ddr p\ddr h.
\end{align*}
\end{proof}

\begin{remark}
Here we present an alternative representation of $p_{f_{v}}(\ell;m,\alpha,\sigma)$ in \eqref{probab_nggp}. It provides a useful tool for implementing a straightforward Monte Carlo evaluation of $p_{f_{v}}(\ell;m,\alpha,\sigma)$. For $\ell=m$,
\begin{align*}
p_{f_{v}}(m;m,\alpha,\sigma)&=\frac{\sigma(\ell-\sigma){m\choose \ell}(1-\sigma)_{(\ell-1)}}{\Gamma(1-\sigma+\ell)}\\
&\quad\times\int_{0}^{+\infty}\int_{0}^{1}\frac{1}{h^{\sigma}}f_{\sigma}(hp)\text{e}^{-h\left(\frac{\alpha2^{-1}}{\sigma}\right)^{\frac{1}{\sigma}}+\frac{\alpha2^{-1}}{\sigma}}p^{m-\ell}(1-p)^{1-\sigma+\ell-1}\ddr p\ddr h\\
&=\frac{1}{\Gamma(m+1)}\int_{0}^{+\infty}\exp\left\{-h\left(\frac{\alpha}{2\sigma}\right)^{1/\sigma}+\frac{\alpha}{2\sigma}\right\}\frac{\sigma\Gamma(m+1)}{\Gamma(m+1-\sigma)}h^{-\sigma}\\
&\quad\times\int_{0}^{1}(1-p)^{m+1-\sigma-1}f_{\sigma}(hp)\ddr p\ddr h\\
&=\frac{(1-\sigma)_{(m)}}{\Gamma(m+1)}\mathbb{E}\left[\exp\left\{-\frac{X}{Y}\left(\frac{\alpha2^{-1}}{\sigma}\right)^{\frac{1}{\sigma}}+\frac{\alpha2^{-1}}{\sigma}\right\}\right],
\end{align*}
where $Y$ is a Beta random variable with parameter $(m-\ell+\sigma,1-\sigma+\ell)$ and $X$ is a random variable, independent of $Y$, distributed according to a polynomially tilted $\sigma$-stable distribution of order $\sigma$, i.e.
\begin{displaymath}
\text{Pr}[X\in\ddr x]=\frac{\Gamma(\sigma+1)}{\Gamma(2)}x^{-\sigma}f_{\sigma}(x)\ddr x.
\end{displaymath}
For $\ell<m$,
\begin{align*}
p_{f_{v}}(\ell;m,\alpha,\sigma)&=\frac{\sigma(\ell-\sigma){m\choose \ell}(1-\sigma)_{(\ell-1)}}{\Gamma(1-\sigma+\ell)}\\
&\quad\times\int_{0}^{+\infty}\int_{0}^{1}\frac{1}{h^{\sigma}}f_{\sigma}(hp)\text{e}^{-h\left(\frac{\alpha2^{-1}}{\sigma}\right)^{\frac{1}{\sigma}}+\frac{\alpha2^{-1}}{\sigma}}p^{m-\ell}(1-p)^{1-\sigma+\ell-1}\ddr p\ddr h\\
&=(\ell-\sigma){m\choose \ell}(1-\sigma)_{(\ell-1)}\\
&\quad\times\frac{\Gamma(m-\ell+\sigma)}{\Gamma(\sigma)\Gamma(m+1)}\\
&\quad\quad\times\int_{0}^{+\infty}f_{\sigma}(hp)\exp\left\{-h\left(\frac{\alpha2^{-1}}{\sigma}\right)^{\frac{1}{\sigma}}+\frac{\alpha2^{-1}}{\sigma}\right\}\frac{\Gamma(\sigma+1)}{\Gamma(2)}h^{-\sigma}\\
&\quad\quad\quad\times\frac{\Gamma(m+1)}{\Gamma(1-\sigma+\ell)\Gamma(m-\ell+\sigma)}\int_{0}^{1}p^{m-\ell}(1-p)^{1-\sigma+\ell-1}\ddr p\ddr h\\
&=\frac{\Gamma(m-\ell+\sigma)}{\Gamma(\sigma)\Gamma(m+1)}\mathbb{E}\left[\exp\left\{-\frac{X}{Y}\left(\frac{\alpha2^{-1}}{\sigma}\right)^{\frac{1}{\sigma}}+\frac{\alpha2^{-1}}{\sigma}\right\}\right].
\end{align*}
According to this alternative representation, $p_{f_{v}}(\ell;m,\alpha,\sigma)$ allows for a Monte Carlo evaluation by sampling from a Beta random variable and from a polynomially tilted $\sigma$-stable random variable of order $\sigma$. See, e.g., \citep{dev09}.
\end{remark}

\textit{Proof of Proposition 2.} The proof follows by a direct application of Lemma \eqref{lem_ngg} by setting $\sigma=1/2$. First, let recall that the density function of the $(1/2)$-stable positive random variable coincides with the IG density function \citep{Ses93} with shape parameter $a=2^{-1/2}$ and scale parameter $b=0$. That is, we write
\begin{displaymath}
f_{1/2}(x)=\frac{1}{2\sqrt{\pi}}w^{-\frac{3}{2}}\exp\left\{-\frac{1}{4w}\right\}.
\end{displaymath}
For $\ell=m$,
\begin{align*}
p_{f_{v}}(m;m,\alpha,\sigma)&=\frac{\alpha2^{m}\left(\frac{1}{2}\right)_{(m)}}{\Gamma(m+1)}\\
&\quad\times\int_{0}^{+\infty}\frac{x^{m}}{(1+2x)^{m+\frac{1}{2}}}\exp\left\{-\alpha2^{1/2}[(2^{-1}+x)^{1/2}-2^{-1/2}]\right\}\de x.
\end{align*}
For $\ell<m$,
\begin{align*}
p_{f_{v}}(\ell;m,\alpha,\sigma)&=\frac{2^{-1}(\ell-2^{-1}){m\choose \ell}\left(\frac{1}{2}\right)_{(\ell-1)}}{\Gamma(2^{-1}+\ell)}\\
&\quad\times\int_{0}^{+\infty}\int_{0}^{1}\frac{1}{\sqrt{h}}f_{1/2}(hp)\text{e}^{-h\alpha^{2}+\alpha}p^{m-\ell}(1-p)^{\frac{1}{2}+\ell-1}\ddr p\ddr h\\
&=(\ell-2^{-1}){m\choose \ell}(1-2^{-1})_{(\ell-1)}\\
&\quad\times\frac{2^{-1}}{2\pi^{1/2}}\text{e}^{\alpha}\int_{0}^{1}\int_{0}^{+\infty}h^{-1-1} \exp\left\{-h\alpha^{2}-\frac{\frac{1}{4p}}{h}\right\}\\
&\quad\quad\times\frac{1}{\Gamma(1-2^{-1}+\ell)}p^{m-\ell-\frac{1}{2}-1}(1-p)^{1-\frac{1}{2}+\ell-1}\ddr p\ddr h\\
&[\text{Equation 3.471.9 of \cite{grad}}]\\
&=(\ell-2^{-1}){m\choose \ell}(1-2^{-1})_{(\ell-1)}\\
&\quad\times\frac{\text{e}^{\alpha}\alpha}{\pi^{1/2}\Gamma(1-2^{-1}+\ell)}\int_{0}^{1}K_{-1}\left(\frac{\alpha}{p^{1/2}}\right)p^{m-\ell-1}(1-p)^{1-\frac{1}{2}+\ell-1}\ddr p,
\end{align*}
where $K_{-1}$ is the modified Bessel function of the second type, or Macdonald function, with parameter $-1$.\qed

\begin{remark}
Here we present an alternative representation of $p_{f_{v}}(\ell;m,\alpha,\sigma)$ in Proposition 2. It provides a useful tool for implementing a straightforward Monte Carlo evaluation of $p_{f_{v}}(\ell;m,\alpha,\sigma)$. For $\ell=m$,
\begin{align*}
p_{f_{v}}(m;m,\alpha,\sigma)&=\frac{\alpha2^{m}\left(\frac{1}{2}\right)_{(m)}}{\Gamma(m+1)}\\
&\quad\times\int_{0}^{+\infty}\frac{x^{m}}{(1+2x)^{m+\frac{1}{2}}}\exp\left\{-\alpha2^{1/2}[(2^{-1}+x)^{1/2}-2^{-1/2}]\right\}\de x\\
&=\frac{1}{\Gamma(m+1)}\int_{0}^{+\infty}\exp\left\{-h\alpha^{2}+\alpha\right\}\frac{2^{-1}\Gamma(m+1)}{\Gamma(m+1-1/2)}\frac{1}{\sqrt{h}}\\
&\quad\times\int_{0}^{1}(1-p)^{m+1-\frac{1}{2}-1}\frac{1}{2\sqrt{\pi}}(hp)^{-\frac{3}{2}}\exp\left\{-\frac{1}{4hp}\right\}\ddr p\ddr h\\
&=\frac{\left(\frac{1}{2}\right)_{(m)}}{\Gamma(m+1)}\mathbb{E}\left[\exp\left\{-\frac{X}{Y}\alpha^{2}+\alpha\right\}\right],
\end{align*}
where $Y$ is a Beta random variable with parameter $(1/2,m+1/2)$ and $X$ is a random variable, independent of $Y$, distributed according to a polynomially tilted IG distribution of the order $1/2$, that is
\begin{displaymath}
\text{Pr}[X\in\ddr x]=\frac{\Gamma(3/2)}{\Gamma(2)}x^{-\frac{1}{2}}\frac{x^{-\frac{3}{2}}}{2\sqrt{\pi}}\exp\left\{-\frac{1}{4x}\right\}\ddr x.
\end{displaymath}
For $\ell<m$,
\begin{align*}
p_{f_{v}}(\ell;m,\alpha,\sigma)&=(\ell-2^{-1}){m\choose \ell}(1-2^{-1})_{(\ell-1)}\\
&\quad\times\frac{\text{e}^{\alpha}\alpha}{\pi^{1/2}\Gamma(1-2^{-1}+\ell)}\int_{0}^{1}K_{-1}\left(\frac{\alpha}{p^{1/2}}\right)p^{m-\ell-1}(1-p)^{1-\frac{1}{2}+\ell-1}\ddr p\\
&=(\ell-2^{-1}){m\choose \ell}(1-2^{-1})_{(\ell-1)}\\
&\quad\times\frac{\Gamma(m-\ell)}{\pi^{1/2}\Gamma(m+1-2^{-1})}\text{e}^{\alpha}\alpha\\
&\quad\quad\quad\times\int_{0}^{1}K_{-1}\left(\frac{\alpha}{p^{1/2}}\right)\frac{\Gamma(m+1-2^{-1})}{\Gamma(1-2^{-1}+\ell)\Gamma(m-\ell)}p^{m-\ell-1}(1-p)^{1-\frac{1}{2}+\ell-1}\ddr p\\
&=(l-2^{-1}){n\choose l}(1-2^{-1})_{(l-1)}\\
&\notag\quad\times\frac{\Gamma(m-\ell)}{\pi^{1/2}\Gamma(m+1-2^{-1})}\text{e}^{\alpha}\alpha\mathbb{E}\left[K_{-1}\left(\frac{\alpha}{Y^{1/2}}\right)\right],
\end{align*}
where $Y$ is a Beta random variable with parameter $(m-\ell,1/2+\ell)$. According to this alternative representation, $p_{f_{v}}(\ell;m,\alpha,\sigma)$ allows for a straightforward Monte Carlo evaluation by sampling from a Beta random variable and from a polynomially tilted IG random variable of order $1/2$. See, e.g., \citep{dev09}.
\end{remark}

\textit{Proof of Theorem 3}.
Because of the assumption of independence of the hash family, we can factorize the marginal likelihood of $(\mathbf{c}_{1},\ldots,\mathbf{c}_{N})$, i.e. of hash functions $h_{1},\ldots,h_{N}$, into the product of the marginal likelihoods of $\mathbf{c}_{n}=(c_{n,1},\ldots,c_{n,J})$, i.e. of each hash function. This, combined with Bayes theorem, leads to 
\begin{align*}
&\text{Pr}[f_v = \ell \mid \{C_{n, h_{n}(v)}\}_{n\in[N]}= \{c_{n, h_{n}(v)}\}_{n\in[N]}] \\
&\quad[\text{Bayes theorem and independence of the hash family}]\\
&\quad=\frac{1}{\text{Pr}[\{C_{n, h_{n}(v)}\}_{n\in[N]}=\{c_{n, h_{n}(v)}\}_{n\in[N]}]}\text{Pr}[f_{v}=\ell]\prod_{n=1}^{N}\text{Pr}[C_{n,h_{n}(v)}=c_{n,h_{n}(v)}\,|\,f_{v}=\ell]\\
&\quad=\frac{1}{\text{Pr}[\{C_{n, h_{n}(v)}\}_{n\in[N]}=\{c_{n, h_{n}(v)}\}_{n\in[N]}]}\text{Pr}[f_{v}=\ell]\prod_{n=1}^{N}\frac{\text{Pr}[C_{n,h_{n}(v)}=c_{n,h_{n}(v)},\,f_{v}=\ell]}{\text{Pr}[f_{v}=\ell]}\\
&\quad=\frac{1}{\text{Pr}[\{C_{n, h_{n}(v)}\}_{n\in[N]}=\{c_{n, h_{n}(v)}\}_{n\in[N]}]}(\text{Pr}[f_{v}=\ell])^{1-N}\prod_{n=1}^{N}\text{Pr}[C_{n,h_{n}(v)}=c_{n,h_{n}(v)}]\text{Pr}[f_{v}=\ell\,|\,C_{n,h_{n}(v)}=c_{n,h_{n}(v)}]\\
&\quad=(\text{Pr}[f_{v}=\ell])^{1-N}\prod_{n=1}^{N}\text{Pr}[f_{v}=\ell\,|\,C_{n,h_{n}(v)}=c_{n,h_{n}(v)}]\\
&\quad\propto\prod_{n=1}^{N}\text{Pr}[f_{v}=\ell\,|\,C_{n,h_{n}(v)}=c_{n,h_{n}(v)}]\\
&\quad[\text{Proposition 2 and Equation 9}]\\
&\quad=\prod_{n\in[N]} \begin{cases} 
\frac{{c_{n,h_n(v)}\choose \ell}e^{\frac{\alpha}{J}}\alpha}{J\pi}\int_{0}^{1}K_{-1}\left(\frac{\alpha}{J\sqrt{x}}\right)x^{c_{n,h_n(v)}-\ell-1}(1-x)^{\frac{1}{2}+\ell-1}\de x&\quad\ell=0,1,\ldots,c_{n,h_n(v)}-1\\[0.4cm] 
\frac{2^{c_{n,h_n(v)}}\alpha\left(\frac{1}{2}\right)_{(c_{n,h_n(v)})}}{J\Gamma(c_{n,h_n(v)}+1)}\int_{0}^{+\infty}\frac{x^{c_{n,h_n(v)}}}{(1+2x)^{c_{n,h_n(v)}+1/2}}e^{-\frac{\alpha}{J}(\sqrt{1+2x}-1)}\de x&\quad \ell=c_{n,h_n(v)},
\end{cases}
\end{align*}
where $K_{-1}(\cdot)$ is the modified Bessel function of the second type, or Macdonald function, with parameter $-1$. 

\section{Estimation of $\alpha$}\label{app_alpha}
We start by deriving the marginal likelihood corresponding to the hashed frequencies $(\mathbf{c}_{1},\ldots,\mathbf{c}_{N})$ induced by the collection of hash functions $h_{1},\ldots,h_{N}$. In particular, according to the definition of $P\sim\text{NIGP}(\alpha,\nu)$ through its family of finite-dimensional distributions, for a single hash function $h_{n}$ the marginal likelihood of $\mathbf{c}_{n}=(c_{n,1},\ldots,c_{n,J})$ is obtained by integrating the normalized IG distribution with parameter $(\alpha/J,\ldots,\alpha/J)$ against the multinomial counts $\mathbf{c}_{n}$. In particular, by means of the normalized IG distribution \eqref{norm_ig}, the marginal likelihood of  $\mathbf{c}_{n}$ has the following expression 
\begin{align*}
&p(\mathbf{c}_{n};\alpha)\\
&\quad=\frac{m!}{\prod_{i=1}^{J}c_{n,i}!}\\
&\quad\quad\times\int_{\{(p_{1},\ldots,p_{J-1})\text{ : }p_{i}\in(0,1)\text{ and } \sum_{i=1}^{J-1}p_{i}\leq 1\}} \prod_{i=1}^{J-1}p_{i}^{c_{n,i}}\left(1-\sum_{i=1}^{J-1}p_{i}\right)^{c_{n,J}}f_{(P_{1},\ldots,P_{J-1})}(p_{1},\ldots,p_{J-1})\ddr p_{1}\cdots\ddr p_{J-1}\\
&\quad=\frac{m!}{\prod_{i=1}^{J}c_{n,i}!}\\
&\quad\quad\times\int_{\{(p_{1},\ldots,p_{J-1})\text{ : }p_{i}\in(0,1)\text{ and } \sum_{i=1}^{J-1}p_{i}\leq 1\}}  \left(\prod_{i=1}^{J}\frac{(\alpha/J)\text{e}^{\alpha/J}}{\sqrt{2\pi}}\right)\prod_{i=1}^{J-1}p_{i}^{c_{n,i}-3/2}\left(1-\sum_{i=1}^{J-1}p_{i}\right)^{c_{n,J}-3/2}\\
&\quad\quad\quad\times\int_{0}^{+\infty}z^{-3J/2+J-1}\exp\left\{-\frac{1}{2z}\left(\sum_{i=1}^{J-1}\frac{(\alpha/J)^{2}}{p_{i}}+\frac{(\alpha/J)^{2}}{1-\sum_{i=1}^{J-1}p_{i}}\right)-\frac{z}{2}\right\}\ddr z\ddr p_{1}\cdots\ddr p_{J-1}\\
&[\text{Change of variable }p_{i}=\frac{x_{i}}{\sum_{i=1}^{k}x_{i}}, \text{ for } i=1,\ldots,J-1, \text{ and } z=\sum_{i=1}^{J}x_{i}]\\
&\quad=\frac{m!}{\prod_{i=1}^{J}c_{n,i}!}\\
&\quad\quad\times\left(\frac{(\alpha/J)\text{e}^{\alpha/J}}{\sqrt{2\pi}}\right)^{J}\int_{(0,+\infty)^{J}}\prod_{i=1}^{J}x_{i}^{c_{n,i}-3/2}\left(\sum_{i=1}^{J}x_{i}\right)^{-\sum_{i=1}^{J}c_{n,i}}\exp\left\{-\frac{1}{2}\sum_{i=1}^{J}\frac{(\alpha/J)^{2}}{x_{i}}-\frac{1}{2}\sum_{i=1}^{J}x_{i}\right\}\ddr x_{1}\cdots\ddr x_{J}\\
&\quad=\frac{m!}{\prod_{i=1}^{J}c_{n,i}!}\\
&\quad\quad\times\left(\frac{(\alpha/J)\text{e}^{\alpha/J}}{\sqrt{2\pi}}\right)^{J}\frac{1}{\Gamma(m)}\int_{(0,+\infty)^{J}}\prod_{i=1}^{J}x_{i}^{c_{n,i}-3/2}\left(\int_{0}^{+\infty}y^{m-1}\exp\left\{-y\sum_{i=1}^{J}x_{i}\right\}\ddr y\right)\\
&\quad\quad\quad\times\exp\left\{-\frac{1}{2}\sum_{i=1}^{J}\frac{(\alpha/J)^{2}}{x_{i}}-\frac{1}{2}\sum_{i=1}^{J}x_{i}\right\}\ddr x_{1}\cdots\ddr x_{J}\\
&\quad=\frac{m!}{\prod_{i=1}^{J}c_{n,i}!}\\
&\quad\quad\times\left(\frac{(\alpha/J)\text{e}^{\alpha/J}}{\sqrt{2\pi}}\right)^{J}\frac{1}{\Gamma(m)}\int_{0}^{+\infty}y^{m-1}\left(\prod_{i=1}^{J}\int_{0}^{+\infty}x_{i}^{c_{n,i}-3/2}\exp\left\{-\frac{(\alpha/J)^{2}}{2x_{i}}-x_{i}\left(y+\frac{1}{2}\right)\right\}\ddr x_{i}\right)\ddr y\\
&[\text{Equation 3.471.9 of \cite{grad}}]\\
&\quad=\frac{m!}{\prod_{i=1}^{J}c_{n,i}!}\\
&\quad\quad\times\left(\frac{(\alpha/J)\text{e}^{\alpha/J}}{\sqrt{2\pi}}\right)^{J}\frac{1}{\Gamma(m)}\int_{0}^{+\infty}y^{m-1}\left(\prod_{i=1}^{J}2\left(\frac{(\alpha/J)^{2}}{1+2y}\right)^{c_{n,i}/2-1/4}K_{c_{n,i}-1/2}\left(\sqrt{c_{n,i}^{2}(1+2y)}\right)\right)\ddr y\\
&\quad=\frac{m\left(\frac{\alpha}{J}\right)^{m+\frac{J}{2}}\text{e}^{\alpha}}{(\pi/2)^{\frac{J}{2}}\prod_{j=1}^{J}c_{n,j}!}\int_{0}^{+\infty}\frac{y^{m-1}}{(1+2y)^{m/2-J/4}}\left(\prod_{i=1}^{J}K_{c_{n,i}-1/2}\left(\sqrt{(\alpha/J)^{2}_{i}(1+2y)}\right)\right)\ddr y.
\end{align*}
Because of the independence of the hash family, $h_{1},\ldots,h_{N}$ leads to the following marginal likelihood of $\{c_{n,j}\}_{n\in[N]\,j\in[J]}$ 
\begin{align}\label{marginale}
&p(\mathbf{c}_{1},\ldots,\mathbf{c}_{N};\alpha)\\
&\notag\quad=\prod_{n\in[N]}\frac{m\left(\frac{\alpha}{J}\right)^{m+\frac{J}{2}}\text{e}^{\alpha}}{(\pi/2)^{\frac{J}{2}}\prod_{j=1}^{J}c_{n,j}!}\int_{0}^{+\infty}\frac{x^{m-1}}{(1+2x)^{\frac{m}{2}-\frac{J}{4}}}\left(\prod_{j=1}^{J}K_{c_{n,j}-\frac{1}{2}}\left(\sqrt{\left(\frac{\alpha}{J}\right)^{2}(1+2x)}\right)\right)\de x.
\end{align}
The marginal likelihood of $\{c_{n,j}\}_{n\in[N]\,j\in[J]}$ in \eqref{marginale} is applied to estimate the mass parameter $\alpha$. This is the empirical Bayes approach to the estimation of $\alpha$. In particular, we consider the following problem
\begin{equation*}
    \argmax_{\alpha}\left\{\prod_{n\in[N]} V_{n,m,\alpha,J} \int_{0}^{+\infty} F_{n,m,\alpha,J}(y) \ddr y\right\},
\end{equation*}
where
\begin{equation*}
    V_{n,m,\alpha,J} = \frac{m\left(\frac{\alpha}{J}\right)^{m+\frac{J}{2}}\text{e}^{\alpha}}{(\pi/2)^{\frac{J}{2}}\prod_{j=1}^{J}c_{n,j}!}
\end{equation*}
and
\begin{equation*}
    F_{n,m,\alpha,J}(y) = \frac{y^{m-1}}{(1+2y)^{\frac{m}{2}-\frac{J}{4}}}\left(\prod_{j=1}^{J}K_{c_{n,j}-\frac{1}{2}}\left(\sqrt{\left(\frac{\alpha}{J}\right)^{2}(1+2y)}\right)\right)
\end{equation*}
under the constraint that $\alpha > 0$. To avoid overflow/underflow issues in the above optimization problem, here we work in log-space. That is, we consider the following equivalent optimization problem\footnote{the computation of log-factorials is done via the specialized implementation of the log-gamma function}
\begin{align*}
    &\argmax_{\alpha}\left\{\sum_{n\in[N]} \log(V_{n,m,\alpha,J}) + \log\left(\int_{0}^{+\infty} F_{n,m,\alpha,J}(y) \ddr y \right)\right\}\\
    &= \argmax_{\alpha}\left\{\sum_{n\in[N]} v_{n,m,\alpha,J} + \log\left(\int_{0}^{+\infty} \exp\{f_{n,m,\alpha,J}(y)\} \ddr y \right)\right\},
\end{align*}
with $v_{n,m,\alpha,J}=\log(V_{n,m,\alpha,J})$ and $f_{n,m,\alpha,J}(y)=\log(F_{n,m,\alpha,J}(y))$. For the computation of the integral we use double exponential quadrature \citep{takahasi1974double}, which approximates $\int_{-1}^{+1} f(y) \ddr y$ with $\sum_{j=1}^{m} w_j f(y_j)$ for appropriate weights $w_j \in \mathcal{W}$ and coordinates $y_j \in \mathcal{Y}$. Integrals of the form $\int_{a}^{b} f(y) \ddr y$ for $-\infty \leq a \leq b \leq +\infty$ are handled via change of variable formulas. To avoid underflow/overflow issues it is necessary to apply the "log-sum-exp" trick to the above integral. That is, 
\begin{align*}
    &\log\left(\int_0^{+\infty} \exp\{f_{n,m,\alpha,J}(y)\} \ddr y\right) = f^* + \log\left(\int_0^{+\infty} \exp\{f_{n,m,\alpha,J}(y) - f^*\} \ddr y\right)
    \end{align*}
    and
\begin{align*}
    &f^* = \argmax_{y\in\mathcal{Y}}\left\{f_{n,m,\alpha,J}(y)\right\}.
\end{align*}
The computation of $\log(K_{c_{n,j}-\frac{1}{2}}(x))$ is performed via the following finite-sum representation of $K_{c_{n,j}-\frac{1}{2}}(x)$, which holds for $K_{v}(x)$ when $v$ is an half-integer. Recall that $K_{v}(x)$ is symmetric in $v$. In particular,
\begin{align*}
&K_{c_{n,i}-1/2}\left(\sqrt{(\alpha/J)(1+2y)}\right)\\
&\quad=\sqrt{\frac{\pi}{2}}\frac{\exp\left\{-((\alpha/J)(1+2y))^{1/2}\right\}}{((\alpha/J)(1+2y))^{1/4}}\sum_{j=0}^{c_{n,i}-1}\frac{(j+c_{n,i}-1)!}{j!(c_{n,i}-j-1)!}(2((\alpha/J)(1+2y))^{1/2})^{-j}.
\end{align*}
In order to increase efficiency in the our optimization, we cache the log-factorials and, anew for each $\alpha$ and $y$ the values of $\log(K_{c_{n,j}-\frac{1}{2}}(\sqrt{(\alpha/J)^{2}(1+2y)}))$ across $j$. In particular, as the dependency on $j$ goes through $c_{n,j}$ only we can exploit the fact that many duplicates exists, i.e. the complexity scales in the number of unique $c_{n,j}$. All code is implemented in \texttt{LuaJIT}\footnote{\href{https://luajit.org}{https://luajit.org}} by using the \texttt{scilua}\footnote{\href{https://scilua.org}{https://scilua.org}} library.

\section{Additional experiments}\label{additional_exp}

We present additional experiments on the application of the CMS-NIGP on synthetic and real data. First, we recall the synthetic and real data to which the CMS-NIGP is applied. As regards synthetic data, we consider datasets of $m=500.000$ tokens from a Zipf's distributions with parameter $s=1.3,\,1.6,\,1.9,\,2,2,\,2.5$. As regards real data, we consider: i) the 20 Newsgroups dataset, which consists of $m=2.765.300$ tokens with $k=53.975$ distinct tokens; ii) the Enron dataset, which consists of $m=6412175$ tokens with $k=28102$ distinct tokens. Tables \ref{tab:experiment_sin_all1}, \ref{tab:experiment_sin_all2}, \ref{tab:experiment_1_full} and \ref{tab:experiment_2_full} report the MAE (mean absolute error) between true frequencies and their corresponding estimates via: i) the CMS-NIGP estimate $\hat{f}_{v}^{\text{\tiny{(NIGP)}}}$; ii) the CMS estimate $\hat{f}_{v}^{\text{\tiny{(CMS)}}}$; iii) the CMS-DP estimate$\hat{f}_{v}^{\text{\tiny{(DP)}}}$, the CMM estimate $\hat{f}_{v}^{\text{\tiny{(CMM)}}}$.

\begin{sidewaystable}[!htb]
    \caption{\scriptsize{Synthetic data: MAE for $\hat{f}_{v}^{\text{\tiny{(NIGP)}}}$, $\hat{f}_{v}^{\text{\tiny{(CMM)}}}$ and $\hat{f}_{v}^{\text{\tiny{(CMS)}}}$, case $J=320,N=2$}}
    \label{tab:experiment_sin_all1}
    \centering
    \begin{tabular}{lrrrrrrrrrrrrrrr}
        \toprule
        \multicolumn{1}{l}{} & \multicolumn{3}{c}{$\mathcal{Z}_{1.3}$} & \multicolumn{3}{c}{$\mathcal{Z}_{1.6}$} & \multicolumn{3}{c}{$\mathcal{Z}_{1.9}$} & \multicolumn{3}{c}{$\mathcal{Z}_{2.2}$} & \multicolumn{3}{c}{$\mathcal{Z}_{2.5}$} \\
        \cmidrule(r){2-4} \cmidrule(r){5-7} \cmidrule(r) {8-10} \cmidrule(r){11-13} \cmidrule(r){14-16}
        Bins $v$ & $\hat{f}_{v}^{\text{\tiny{(CMS)}}}$ & $\hat{f}_{v}^{\text{\tiny{(CMM)}}}$ & $\hat{f}_{v}^{\text{\tiny{(NIGP)}}}$ & $\hat{f}_{v}^{\text{\tiny{(CMS)}}}$ & $\hat{f}_{v}^{\text{\tiny{(CMM)}}}$ & $\hat{f}_{v}^{\text{\tiny{(NIGP)}}}$ & $\hat{f}_{v}^{\text{\tiny{(CMS)}}}$ & $\hat{f}_{v}^{\text{\tiny{(CMM)}}}$ & $\hat{f}_{v}^{\text{\tiny{(NIGP)}}}$ & $\hat{f}_{v}^{\text{\tiny{(CMS)}}}$ & $\hat{f}_{v}^{\text{\tiny{(CMM)}}}$ & $\hat{f}_{v}^{\text{\tiny{(NIGP)}}}$ & $\hat{f}_{v}^{\text{\tiny{(CMS)}}}$ & $\hat{f}_{v}^{\text{\tiny{(CMM)}}}$ & $\hat{f}_{v}^{\text{\tiny{(NIGP)}}}$ \\[0.05cm]
        \midrule
            (0,1] &  1,061.3 & 161.72 & 231.31 &    629.40 &   62.19 &  134.75 &    308.11 &   81.10 &  65.71 &  51.65 &   1.04 &  12.91 & 32.65 &   1.02 &   7.16 \\
            (1,2] &  1,197.9 & 169.74 & 287.43 &    514.31 &  102.42 &  119.22 &    154.20 &    2.00 &  37.03 & 289.50 &   2.04 &  61.87 & 48.15 &   2.01 &   9.88 \\
            (2,4] &  1,108.3 & 116.37 & 262.18 &    474.82 &   52.10 &   95.78 &  2,419.51 & 2215.85 & 353.73 & 134.05 &   3.40 &  26.90 & 54.34 &  10.50 &  10.09 \\
            (4,8] &  1,275.9 & 378.04 & 302.89 &    786.73 &  214.46 &  175.10 &    460.13 &  258.90 &  83.30 & 118.40 &   6.44 &  21.58 & 69.85 &   6.03 &  14.28 \\
           (8,16] &  1,236.1 & 230.32 & 257.08 &    719.84 &  232.24 &  136.66 &    380.05 &  139.50 &  66.44 & 413.13 & 129.03 &  77.39 & 80.80 &  13.10 &  20.15 \\
          (16,32] &  1,256.8 & 221.98 & 248.41 &    831.70 &   79.73 &  190.05 &    288.59 &   23.90 &  41.99 & 503.60 & 364.30 &  90.29 &  9.86 &  22.39 &  15.36 \\
          (32,64] &  1,312.8 & 235.87 & 284.12 &    783.90 &  184.99 &  139.52 &    415.58 &   54.82 &  67.30 & 217.81 &  82.92 &  48.00 & 10.22 &  30.90 &  28.90 \\
         (64,128] &  1,721.7 & 766.29 & 312.59 &    950.31 &  304.36 &  125.07 &  1,875.50 & 1762.20 & 353.10 &  64.01 &  97.40 &  65.91 & 13.75 &  96.98 &  66.18 \\
        (128,256] &  1,107.7 & 334.57 &  97.91 &  1,727.19 & 1488.38 &  273.50 &    202.09 &  163.61 & 110.32 &  46.80 & 156.71 & 130.94 & 17.51 & 181.38 & 125.75 \\
        \bottomrule
    \end{tabular}
    \end{sidewaystable}

\begin{sidewaystable}[!htb]
    \caption{\scriptsize{Synthetic data: MAE for $\hat{f}_{v}^{\text{\tiny{(NIGP)}}}$, $\hat{f}_{v}^{\text{\tiny{(CMM)}}}$ and $\hat{f}_{v}^{\text{\tiny{(CMS)}}}$, case $J=160,N=4$}}
    \label{tab:experiment_sin_all2}
    \centering
    \begin{tabular}{lrrrrrrrrrrrrrrr}
        \toprule
        \multicolumn{1}{l}{} & \multicolumn{3}{c}{$\mathcal{Z}_{1.3}$} & \multicolumn{3}{c}{$\mathcal{Z}_{1.6}$} & \multicolumn{3}{c}{$\mathcal{Z}_{1.9}$} & \multicolumn{3}{c}{$\mathcal{Z}_{2.2}$} & \multicolumn{3}{c}{$\mathcal{Z}_{2.5}$} \\
        \cmidrule(r){2-4} \cmidrule(r){5-7} \cmidrule(r) {8-10} \cmidrule(r){11-13} \cmidrule(r){14-16}
        Bins $v$ & $\hat{f}_{v}^{\text{\tiny{(CMS)}}}$ & $\hat{f}_{v}^{\text{\tiny{(CMM)}}}$ & $\hat{f}_{v}^{\text{\tiny{(NIGP)}}}$ & $\hat{f}_{v}^{\text{\tiny{(CMS)}}}$ & $\hat{f}_{v}^{\text{\tiny{(CMM)}}}$ & $\hat{f}_{v}^{\text{\tiny{(NIGP)}}}$ & $\hat{f}_{v}^{\text{\tiny{(CMS)}}}$ & $\hat{f}_{v}^{\text{\tiny{(CMM)}}}$ & $\hat{f}_{v}^{\text{\tiny{(NIGP)}}}$ & $\hat{f}_{v}^{\text{\tiny{(CMS)}}}$ & $\hat{f}_{v}^{\text{\tiny{(CMM)}}}$ & $\hat{f}_{v}^{\text{\tiny{(NIGP)}}}$ & $\hat{f}_{v}^{\text{\tiny{(CMS)}}}$ & $\hat{f}_{v}^{\text{\tiny{(CMM)}}}$ & $\hat{f}_{v}^{\text{\tiny{(NIGP)}}}$ \\[0.05cm]
        \midrule
            (0,1] &  212.1 & 590.48 &   0.94 &  262.00 & 146.11 &   0.25 &  424.8 & 130.90 &   0.18 & 154.79 &  47.10 &   0.32 & 56.7 &   1.01 &   0.38 \\
            (1,2] &  339.8 & 359.57 &   0.56 &  332.75 &  63.21 &   0.70 &  552.0 &  65.00 &   0.82 & 182.72 &   2.01 &   1.24 & 48.2 &   2.03 &   1.45 \\
            (2,4] &  270.9 &  69.42 &   1.33 &  277.80 & 301.89 &   2.47 &  487.3 & 163.55 &   2.53 & 184.70 &  97.15 &   2.66 & 57.8 &  14.35 &   2.74 \\
            (4,8] &  234.6 & 339.95 &   4.69 &  375.74 & 579.94 &   4.67 &  545.2 & 243.08 &   5.28 & 252.53 &  62.70 &   5.96 & 51.1 &   8.30 &   5.42 \\
           (8,16] &  213.3 & 313.37 &  10.57 &  165.73 & 152.53 &  10.68 &  493.2 & 196.20 &  10.86 & 247.33 &  29.70 &  10.28 & 24.1 &  14.11 &  11.75 \\
          (16,32] &  283.0 &  23.30 &  20.72 &  217.20 &  22.94 &  19.21 &  535.5 & 154.30 &  22.08 & 295.90 & 190.92 &  21.57 & 25.0 &  23.20 &  23.37 \\
          (32,64] &  305.7 & 133.09 &  42.66 &  284.61 & 209.13 &  43.14 &  637.8 & 150.05 &  42.64 & 120.62 &  71.86 &  44.49 & 31.7 &  40.40 &  44.03 \\
         (64,128] &  244.5 & 102.43 &  92.26 &  120.21 & 118.42 &  94.43 &  425.1 & 198.60 &  95.19 & 180.30 & 113.75 &  95.10 & 29.2 &  94.73 &  93.34 \\
        (128,256] &  237.4 & 294.43 & 170.09 &  141.30 & 573.12 & 173.87 &  525.9 & 267.15 & 185.83 & 129.70 & 176.50 & 180.41 & 32.1 & 119.19 & 179.51 \\
        \bottomrule
    \end{tabular}
\end{sidewaystable}

\begin{table}[!htb]
    \caption{\scriptsize{Real data ($J=12000$ and $N=2$): MAE for $\hat{f}_{v}^{\text{\tiny{(NIGP)}}}$, $\hat{f}_{v}^{\text{\tiny{(DP)}}}$ and $\hat{f}_{v}^{\text{\tiny{(CMS)}}}$}}
    \label{tab:experiment_1_full}
    \centering
    \resizebox{0.8\textwidth}{!}{    \begin{tabular}{lrrrrrr}
        \toprule
        \multicolumn{1}{l}{} & \multicolumn{3}{c}{20 Newsgroups} & \multicolumn{3}{c}{Enron} \\
        \cmidrule(r){2-4} \cmidrule(r){5-7}
        Bins for true $v$ & $\hat{f}_{v}^{\text{\tiny{(CMS)}}}$ &  $\hat{f}_{v}^{\text{\tiny{(DP)}}}$ &  $\hat{f}_{v}^{\text{\tiny{(NIGP)}}}$ & $\hat{f}_{v}^{\text{\tiny{(CMS)}}}$ & $\hat{f}_{v}^{\text{\tiny{(DP)}}}$ &  $\hat{f}_{v}^{\text{\tiny{(NIGP)}}}$ \\
        \midrule
                 (0,1] &  46.4 &      46.39 &        11.34 &   12.2 &      12.20 &         3.00 \\
                 (1,2] &  16.6 &      16.60 &         3.53 &   13.8 &      13.80 &         3.06 \\
                 (2,4] &  38.4 &      38.40 &         7.71 &   61.5 &      61.49 &        12.55 \\
                 (4,8] &  59.4 &      59.39 &        10.40 &   88.4 &      88.39 &        17.36 \\
                (8,16] &  54.3 &      54.29 &        11.34 &   23.4 &      23.40 &         4.58 \\
               (16,32] &  17.8 &      17.80 &         9.85 &   55.1 &      55.09 &        11.58 \\
               (32,64] &  40.8 &      40.79 &        25.65 &  128.5 &     128.48 &        39.46 \\
              (64,128] &  26.0 &      25.99 &        57.95 &  131.1 &     131.08 &        54.42 \\
             (128,256] &  13.6 &      13.59 &       126.07 &   50.7 &      50.68 &       119.04 \\
        \bottomrule
    \end{tabular}}
\end{table}

\begin{table}[!htb]
    \caption{\scriptsize{Real data ($J=8000$ and $N=4$): MAE for $\hat{f}_{v}^{\text{\tiny{(NIGP)}}}$, $\hat{f}_{v}^{\text{\tiny{(DP)}}}$ and $\hat{f}_{v}^{\text{\tiny{(CMS)}}}$}}
    \label{tab:experiment_2_full}
    \centering
    \resizebox{0.8\textwidth}{!}{    \begin{tabular}{lrrrrrr}
        \toprule
        \multicolumn{1}{l}{} & \multicolumn{3}{c}{20 Newsgroups} & \multicolumn{3}{c}{Enron} \\
        \cmidrule(r){2-4} \cmidrule(r){5-7}
        Bins for true $v$ & $\hat{f}_{v}^{\text{\tiny{(CMS)}}}$ &  $\hat{f}_{v}^{\text{\tiny{(DP)}}}$ &  $\hat{f}_{v}^{\text{\tiny{(NIGP)}}}$ & $\hat{f}_{v}^{\text{\tiny{(CMS)}}}$ & $\hat{f}_{v}^{\text{\tiny{(DP)}}}$ &  $\hat{f}_{v}^{\text{\tiny{(NIGP)}}}$ \\
        \midrule
                 (0,1] &  53.4 &      53.39 &         0.39 &   71.0 &      70.98 &         0.41 \\
                 (1,2] &  30.5 &      30.49 &         1.40 &   47.4 &      47.38 &         1.47 \\
                 (2,4] &  32.5 &      32.49 &         2.70 &   52.5 &      52.49 &         3.25 \\
                 (4,8] &  38.7 &      38.69 &         5.97 &   53.1 &      53.08 &         6.17 \\
                (8,16] &  25.3 &      25.29 &        11.97 &   57.0 &      56.98 &        11.28 \\
               (16,32] &  25.0 &      24.99 &        21.25 &   90.0 &      89.98 &        19.82 \\
               (32,64] &  39.7 &      39.69 &        42.81 &  108.4 &     108.37 &        47.07 \\
              (64,128] &  22.1 &      22.09 &        91.06 &   55.7 &      55.67 &        87.32 \\
             (128,256] &  25.8 &      25.79 &       205.58 &   80.8 &      80.76 &       178.23 \\
        \bottomrule
    \end{tabular}}
\end{table}

\bibliographystyle{apalike}
\bibliography{ref}